\documentclass[twoside]{article}
\usepackage{ecj,palatino,epsfig,latexsym,natbib}

\usepackage{amssymb}
\usepackage{amsmath}
\usepackage{graphicx}
\usepackage[utf8]{inputenc}
\usepackage{color}
\usepackage{algorithm}
\usepackage[noend]{algorithmic}

\usepackage{wasysym} 
\usepackage{xstring}
\usepackage{graphicx}
\usepackage{float}
\usepackage{rotating}
\usepackage{lscape}
\usepackage{colortbl}
\usepackage[table]{xcolor}
\usepackage{url}

\usepackage{amssymb,amsmath}
\usepackage{graphicx}
\usepackage{xcolor}
\usepackage{float}
\usepackage{rotating}
\usepackage{hyperref}

\newcommand{\ma}[1]{\mathchoice{\mbox{\boldmath$\displaystyle#1$}}
  {\mbox{\boldmath$\textstyle#1$}} {\mbox{\boldmath$\scriptstyle#1$}}
  {\mbox{\boldmath$\scriptscriptstyle#1$}}}
\renewcommand{\ma}[1]{\mathnormal{\mathbf{#1}}}
\newcommand{\mstr}[1]{\mathrm{#1}}
\newcommand{\C}{ \ensuremath{\ma{C}} }

\newcommand{\dd}{n}
\newcommand{\R}{\mathbb{R}}
\newcommand{\vc}[1]{\textit{\textbf{#1}}}
\def\ONE{{\rm 1\hspace{-0.50ex}1}}

\def\RR{{\rm I\hspace{-0.50ex}R}}

\newcommand{\A}{ \ensuremath{\ma{A}} }

\newcommand{\I}{ \ensuremath{\ma{I}} }

\def\NormOI{{\mathcal N}  \hspace{-0.13em}\left({\ma{0}, \ensuremath{\ma{I}}\,}\right)}
\def\ONE{{\rm 1\hspace{-0.80ex}1}}
\def\Id{\ensuremath{\ma{I}}}

\def\x{\vc{x}}

\def\m{\vc{m}}

\begin{document}

\ecjHeader{x}{x}{xxx-xxx}{201X}{LM-CMA for Large Scale Black-box Optimization}{I. Loshchilov}
\title{\bf  LM-CMA: an Alternative to L-BFGS for \\Large Scale Black-box Optimization}  

\author{\name{\bf Ilya Loshchilov} \hfill \addr{ilya.loshchilov@gmail.com}\\ 
        \addr{ Laboratory of Intelligent Systems (LIS)\\ \'Ecole Polytechnique F\'ed\'eral de Lausanne (EPFL), Lausanne, Switzerland}
}

\maketitle

\begin{abstract}

The limited memory BFGS method (L-BFGS) of Liu and Nocedal (1989) is often considered to be the method of choice for continuous optimization when first- and/or second- order information is available. However, the use of L-BFGS can be complicated in a black-box scenario where gradient information is not available and therefore should be numerically estimated. The accuracy of this estimation, obtained by finite difference methods,  is often problem-dependent that may lead to premature convergence of the algorithm. 

In this paper, we demonstrate an alternative to L-BFGS, the limited memory Covariance Matrix Adaptation Evolution Strategy (LM-CMA) proposed by Loshchilov (2014). The LM-CMA is a stochastic derivative-free algorithm for numerical optimization of non-linear, non-convex optimization problems. Inspired by the L-BFGS, the LM-CMA samples candidate solutions according to a covariance matrix reproduced from $m$ direction vectors selected during the optimization process. The decomposition of the covariance matrix into Cholesky factors allows to reduce the memory complexity to $O(mn)$, where $n$ is the number of decision variables. The time complexity of sampling one candidate solution is also $O(mn)$, but scales as only about 25 scalar-vector multiplications in practice. The algorithm has an important property of invariance w.r.t. strictly increasing transformations of the objective function, such transformations do not compromise its ability to approach the optimum. The LM-CMA outperforms the original CMA-ES and its large scale versions on non-separable ill-conditioned problems with a factor increasing with problem dimension. Invariance properties of the algorithm do not prevent it from demonstrating a comparable performance to L-BFGS on non-trivial large scale smooth and nonsmooth optimization problems.

\end{abstract}


\section{Introduction}
In a black-box scenario, knowledge about an objective function $f: \vc{X} \rightarrow \R$, to be optimized on some space $\vc{X}$, is restricted to the handling of a device that delivers the value of $f(\vc{x})$ for any input $\vc{x} \in \vc{X}$. The goal of black-box optimization is to find solutions with small (in the case of minimization) value $f(\vc{x})$, using the least number of calls to the function $f$ \citep{2011OllivierIGO}. In continuous domain, $f$ is defined as a mapping $\R^n \rightarrow \R$, where $n$ is the number of variables. 
The increasing typical number of variables involved in everyday optimization problems makes it harder to supply the search with  useful problem-specific knowledge, e.g., gradient information, valid assumptions about problem properties. 
The use of large scale black-box optimization approaches would seem attractive providing that a comparable 
performance can be achieved. 

The use of well recognized gradient-based approaches such as the Broyden-Fletcher-Goldfarb-Shanno (BFGS) algorithm \citep{1970ShannoBFGS} is complicated in the black-box scenario since gradient information is not available and therefore should be estimated by costly finite difference methods (e.g., $n+1$ function evaluations per gradient estimation for forward difference and $2n+1$ for central difference). The latter procedures are problem-sensitive and may require a priori knowledge about the problem at hand, e.g., scaling of $f$, decision variables and expected condition number \citep{li2007effective}. 

By the 1980s, another difficulty has become evident: the use of quasi-Newton methods such as BFGS is limited to small and medium scale optimization problems for which the approximate inverse Hessian matrix can be stored in memory. As a solution, it was proposed not to store the matrix but to reconstruct it using information from the last $m$ iterations \citep{nocedal1980updating}. The final algorithm called the limited memory BFGS algorithm (L-BFGS or LM-BFGS) proposed by \cite{liu1989limited} is still considered to be the state-of-the-art of large scale gradient-based optimization \citep{becker2012quasi}. 
However, when a large scale black-box function is considered, the L-BFGS is forced to deal both with a scarce information coming from only $m$ recent gradients and potentially numerically imprecise estimations of these gradients which scale up the run-time in the number of function evaluations by a factor of $n$. It is reasonable to wonder whether the L-BFGS and other derivative-based algorithms are still competitive in these settings or better performance and robustness can be achieved by derivative-free algorithms.

The Covariance Matrix Adaptation Evolution Strategy (CMA-ES) seems to be a reasonable alternative, it is a derivative-free algorithm designed to learn dependencies between decision variables by adapting a covariance matrix which defines the sampling distribution of candidate solutions \citep{2003HansenCMA}. This algorithm constantly demonstrates good performance at various platforms for comparing continuous optimizers such as the Black-Box Optimization Benchmarking (BBOB) workshop \citep{2009FinckBBOB2009setup,2010HansenBBOBAllTables,2013LoshchilovHCMA} and the Special Session at Congress on Evolutionary Computation \citep{2009GarciaCEC2005results,2013LoshchilovCEC}. The CMA-ES was also extended to noisy \citep{hansen2009method}, expensive \citep{2006KernlmmCMA,loshchilov2012self} and multi-objective optimization \citep{igel2007covariance}.

The principle advantage of CMA-ES, the learning of dependencies between $n$ decision variables, also forms its main practical limitations such as $O(n^2)$ memory storage required to run the algorithm and $O(n^2)$ computational time complexity per function evaluation \citep{ros2008simple}. These limitations may preclude the use of CMA-ES for computationally cheap but large scale optimization problems if the internal computational cost of CMA-ES is greater than the cost of one function evaluation. On non-trivial large scale problems with $n>10,000$ not only the internal computational cost of CMA-ES becomes enormous but it is becoming simply impossible to efficiently store the covariance matrix in memory. 
An open problem is how to extend efficient black-box approaches such as CMA-ES to $n\gg1000$ while keeping a reasonable trade-off between the performance in terms of the number of function evaluations and the internal time and space complexity. The low complexity methods such as separable CMA-ES (sep-CMA-ES by \cite{ros2008simple}), linear time Natural Evolution Strategy (R1-NES by \cite{sun2011linear}) and VD-CMA by \cite{akimoto2014comparison} are useful when the large scale optimization problem at hand is separable or decision variables are weakly correlated, otherwise the performance of these algorithms w.r.t. the original CMA-ES may deteriorate significantly. 

In this paper, we present a greatly improved version of the recently proposed extension of CMA-ES to large scale optimization called the limited memory CMA-ES (LM-CMA) by \cite{2014LoshchilovLMCMA}. Instead of storing the covariance matrix, the LM-CMA stores $m$ direction vectors in memory and uses them to generate solutions. 
The algorithm has $O(mn)$ space complexity, where $m$ can be a function of $n$. 
The time complexity is linear in practice with the smallest constant factor among the presented evolutionary algorithms.

The paper is organized as follows. First, we briefly describe L-BFGS in Section \ref{lbfgssection} and CMA-ES with its large scale alternatives in Section \ref{state}. Then, we present the improved version of LM-CMA in Section \ref{lmcmasection}, investigate its performance w.r.t. large scale alternatives in Section \ref{resultssection} and conclude the paper in Section \ref{conclusion}.

\section{The L-BFGS}
\label{lbfgssection}

An early version of the L-BFGS method, at that time called the SQN method, was proposed by \cite{nocedal1980updating}.
During the first $m$ iterations the L-BFGS is identical to the BFGS method, but stores BFGS corrections separately until the maximum number of them $m$ is used up. Then, the oldest corrections are replaced by the newest ones. The approximate of the inverse Hessian of $f$ at iteration $k$, $\vc{H}_k$ is obtained by applying $m$ BFGS updates to a sparse symmetric and positive definite matrix $\vc{H}_0$ provided by the user \citep{liu1989limited}.

Let us denote iterates by $\vc{x}_k$, $\vc{s}_k=\vc{x}_{k+1}-\vc{x}_k$ and $\vc{y}_k=\vc{g}_{k+1}-\vc{g}_k$, where $\vc{g}$ denotes gradient. 
The method uses the inverse BFGS formula in the form 

	\begin{equation}
  \vc{H}_{k+1} = \vc{V}_k^T \vc{H}_k \vc{V}_k + \rho_k \vc{s}_k \vc{s}^T_k,
  \end{equation}

where $\rho_k=1/\vc{y}^T_k \vc{s}_k$, and 
$\vc{V}_k = \vc{I} - \rho_k \vc{y}_k \vc{s}_k^T$ \citep{dennis1996numerical,liu1989limited}.

The L-BFGS method works as follows \citep{liu1989limited}:

\textit{Step 1}. Choose $\vc{x}_0$, $m$, $0<\beta'<1/2$, 
$\beta'<\beta<1$, and a symmetric and positive definite starting matrix $\vc{H}_0$. Set $k=0$.

\textit{Step 2}. Compute 

	\begin{equation}
  \vc{d}_{k} = -\vc{H}_k \vc{g}_k,
  \end{equation}
	\begin{equation}
  \vc{x}_{k+1} = \vc{x}_k + \alpha_k \vc{d}_k,
  \end{equation}
	
	where $\alpha_k$ satisfies the Wolfe conditions \citep{wolfe1969convergence}:
	
	\begin{equation}
  f(\vc{x}_k + \alpha_k \vc{d}_k) \leq f(\vc{x}_k) + \beta' \alpha_k \vc{g}_k^T \vc{d}_k,
  \end{equation}
	
	\begin{equation}
 g(\vc{x}_k + \alpha_k \vc{d}_k)^T\vc{d}_k \geq \beta \vc{g}_k^T \vc{d}_k.
  \end{equation}

The novelty introduced by \cite{liu1989limited} w.r.t. the version given in \cite{nocedal1980updating} is that the line search is not forced to perform at least one cubic 
interpolation, but the unit steplength $\alpha_k=1$ is always tried first, and if it satisfies the Wolfe conditions, it is accepted. 
\newpage
\textit{Step 3}. Let $\hat{m}=\min(k,m-1)$. Update $\vc{H}_0$ $\hat{m}+1$ times using the pairs $\left\{\vc{y}_j,\vc{s}_j \right\}^k_{j=k-\hat{m}}$ as follows:

\[ \begin{array}{lll}
\lefteqn{ \vc{H}_{k+1} = (\vc{V}^T_k \cdot \cdot \cdot \vc{V}^T_{k-\hat{m}}) \vc{H}_0 (\vc{V}_{k-\hat{m}} \cdot \cdot \cdot \vc{V}_{k})}\\
 && + \rho_{k-\hat{m}}  (\vc{V}^T_k \cdot \cdot \cdot \vc{V}^T_{k-\hat{m}+1}) \vc{s}_{k-\hat{m}} \vc{s}^T_{k-\hat{m}} (\vc{V}_{k-\hat{m}+1}  \cdot \cdot \cdot \vc{V}_k)\\
 && + \rho_{k-\hat{m}+1}  (\vc{V}^T_k \cdot \cdot \cdot \vc{V}^T_{k-\hat{m}+2}) \vc{s}_{k-\hat{m}+1} \vc{s}^T_{k-\hat{m}+1} (\vc{V}_{k-\hat{m}+2}  \cdot \cdot \cdot \vc{V}_k)\\
&& \vdots \\
&& + \rho_k \vc{s}_k \vc{s}_k^T
\end{array} \]

\textit{Step 4}. Set $k=k+1$ and go to Step 2.

The algorithm space and time complexity scales as $O(mn)$ per iteration (not per function evaluation), where $m$ in order of 5-40 suggested in the original paper is still the most common setting. An extension to bound constrained optimization called L-BFGS-B has the efficiency of the original algorithm,  however at the cost of a significantly more complex implementation \citep{byrd1995limited}. Extensions to optimization with arbitrary constraints are currently not available. Satisfactory and computationally tractable handling of noise is at least problematic, often impossible. 

Nevertheless, as already mentioned above, when gradient information is available, L-BFGS is competitive to other techniques \citep{becker2012quasi,ngiam2011optimization} and often can be viewed as a method of choice \citep{andrew2007scalable} for large scale continuous optimization. However, in the black-box scenario when gradient-information is not available (direct search settings), the advantages of L-BFGS are becoming less obvious and derivative-free algorithms can potentially perform comparable. In this paper, we investigate this scenario in detail.

\section{Evolution Strategies for Large Scale Optimization}
\label{state}

Historically, first Evolution Strategies \citep{1973RechenbergEvolutionsstrategie} were designed to perform the search without learning dependencies between variables which is a more recent development that gradually led to the CMA-ES algorithm \citep{1996HansenCMAES,2003HansenCMA}.
In this section, we discuss the CMA-ES algorithm and its state-of-the-art derivatives for large scale optimization. For a recent comprehensible overview of Evolution Strategies, the interested reader is referred to \cite{hansen2015evolution}. 
More specifically, the analysis of theoretical foundations of Evolution Strategies is provided by \cite{wierstra2014natural,2011OllivierIGO,2013Akimoto,glasmachers2012convergence,auger2013linear,hansen2014principled,arnold2014behaviour,beyerconvergence}.

\subsection{The CMA-ES}
\label{sectionCMA}

\begin{algorithm}[tb!]
\caption{The ($\mu/\mu_{w},\lambda$)-CMA-ES}
\label{CMAdefault}
\begin{algorithmic}[1]
\STATE{\textbf{given} $n \in \mathbb{N}_+$, $\lambda = 4 + \lfloor 3 \ln \, n  \rfloor $, $\mu =  \lfloor \lambda/2   \rfloor $, 
											$\vc{w}_i = \frac{ \ln(\mu + \frac{1}{2}) - \ln\,i}{ \sum^{\mu}_{j=1}(\ln(\mu + \frac{1}{2})-\ln\,j)} \; \mstr{for} \; i=1 \ldots \mu$,
											$\mu_w = \frac{1}{\sum^{\mu}_{i=1} w^2_i}$,
											$c_{\sigma} = \frac{\mu_w + 2}{n+\mu_w+3}$,      $d_{\sigma} = 1 + c_{\sigma} +2 \, \mstr{max}(0,\sqrt{ \frac{\mu_w - 1}{n+1}}-1)$,
											$c_c = \frac{4}{n+4}$, $c_1 = \frac{2 \, \mstr{\min}(1,\lambda/6)}{(n+1.3)^2 +\mu_w}$, $c_{\mu} = \frac{2 \, (\mu_w -2 + 1/{\mu_w})}{(n+2)^2+\mu_w}$} \label{CMAEScmaGiven}
\STATE{\textbf{initialize} $\vc{m}^{t=0} \in \R^{\dd}, \sigma^{t=0} > 0, \vc{p}^{t=0}_{\sigma} = \ma{0}, \vc{p}^{t=0}_{c} = \ma{0}, \C^{t=0} = \Id, t \leftarrow 0 $}
\REPEAT
  \FOR{$k \leftarrow 1,\ldots,\lambda$} \label{CMAGenerateBegin}
			\STATE{ $\vc{x}_k \leftarrow \vc{m}^t + \sigma^{t}  {{\mathcal N}  \hspace{-0.13em}\left({\ma{0},\C^{t}\,}\right)} $} \label{CMAsampling}
			\STATE{ $\vc{f}_k \leftarrow f(\vc{x}_k)$} \label{CMAGenerateEnd}
  \ENDFOR
	\STATE{ $ \vc{m}^{t+1} \leftarrow \sum_{i=1}^{\mu} \vc{w}_i \vc{x}_{i:\lambda} \;$} // the symbol $i:\lambda$ denotes $i$-th best individual on $f$ \label{CMAComputeNewMean}
	\STATE{ $ \vc{p}^{t+1}_{\sigma} \leftarrow (1 - c_{\sigma}) \vc{p}^{t}_{\sigma} + \sqrt{c_{\sigma}(2-c_{\sigma})} \sqrt{\mu_w} {\C^t}^{-\frac{1}{2}} \frac{\vc{m}^{t+1}-\vc{m}^{t}}{\sigma^{t}} $} \label{CMASigmaPathUpdate}
	\STATE{ $ h_{\sigma} \leftarrow \ONE_{ \left\| p^{t+1}_{\sigma} \right\| < \sqrt{1 - (1-c_{\sigma})^{2(t+1)}}(1.4 + \frac{2}{n+1}) \, \mathbb{E} \left\| \NormOI \right\|  } $} \label{CMAhsigma}
	\STATE{ $ \vc{p}^{t+1}_{c} \leftarrow (1 - c_{c}) \vc{p}^{t}_{c} + h_{\sigma} \sqrt{c_{c}(2-c_{c})} \sqrt{\mu_w} \frac{\vc{m}^{t+1}-\vc{m}^{t}}{\sigma^{t}} $} \label{CMAEvoPathUpdate}
	\STATE{ $ \C_{\mu} \leftarrow \sum^{\mu}_{i=1} w_i \frac{\x_{i:\lambda} - \m^t}{\sigma^t} \times \frac{(\x_{i:\lambda} - \m^t)^T}{\sigma^t}$ }  \label{CMAplusCov}
	\STATE{ $ \C^{t+1} \leftarrow (1 - c_1 - c_{\mu}) \C^t	+ 
						c_1 \underbrace{\vc{p}^{t+1}_c {\vc{p}^{t+1}_c}^T}_{\mstr{rank-one\,update}} + 
						c_{\mu} \hspace{-1.9em} \underbrace{\C_{\mu}}_{\mstr{rank-\mu \,update}}$}
						\label{CMAupdate}
	\STATE{ $ \sigma^{t+1} \leftarrow \sigma^{t} \mstr{exp}
	          ( \frac{c_{\sigma}}{d_{\sigma}} (  \frac{\left\| \vc{p}^{t+1}_{\sigma} \right\|}{ \mathbb{E} \left\| \NormOI \right\| } - 1  )) $} \label{CMAStepSizeUpdate}
  \STATE{ $ t \leftarrow t + 1$}
\UNTIL{ \textit{stopping criterion is met} }
\end{algorithmic}
\end{algorithm}

The Covariance Matrix Adaptation Evolution Strategy \citep{1996HansenCMAES,2001HansenCMA,2003HansenCMA} is probably the most popular and in overall the most efficient Evolution Strategy. 

The ($\mu/\mu_{w},\lambda$)-CMA-ES is outlined in \textbf{Algorithm} \ref{CMAdefault}. 
At iteration $t$ of CMA-ES, a mean $\vc{m}^t$ of the mutation distribution (can be interpreted as an estimation of the optimum) is used to generate its $k$-th out of $\lambda$ candidate solution $\vc{x}_k \in \mathbb{R}^\dd$ 
	 (line \ref{CMAsampling}) by adding a random Gaussian mutation defined by a (positive definite) covariance matrix \newpage $\C^t \in \R^{n \times n}$ and a mutation step-size $\sigma^t$ as follows:
	
	\begin{equation}
  \vc{x}^t_k \leftarrow {\mathcal N}  \hspace{-0.13em}\left({\vc{m}^t,{\sigma^t}^2 {\C}^t}\right) \leftarrow \vc{m}^t + \sigma^t {\mathcal N}  \hspace{-0.13em}\left({\ma{0},{\C}^t}\right)
  \end{equation}
	
	These $\lambda$ solutions then should be evaluated with an objective function $f$ (line \ref{CMAGenerateEnd}).
	The old mean of the mutation distribution is stored in $\vc{m}^{t}$ and a new mean $\vc{m}^{t+1}$ is computed as a \textit{weighted sum} 
	of the best $\mu$ parent individuals selected among $\lambda$ generated offspring individuals (line \ref{CMAComputeNewMean}).
	The weights $\vc{w}$ are used to control the impact of the selected individuals, the weights are usually higher for better ranked individuals (line \ref{CMAEScmaGiven}).
	
	The procedure of the adaptation of the step-size $\sigma^t$ in CMA-ES is inherited from the Cumulative Step-Size Adaptation Evolution Strategy (CSA-ES) \citep{1996HansenCMAES} and is controlled by evolution path $\vc{p}_{\sigma}^{t+1}$.
	Successful mutation steps $\frac{\vc{m}^{t+1}-\vc{m}^{t}}{\sigma^{t}}$ (line \ref{CMASigmaPathUpdate}) are tracked in the space of sampling, i.e., in the isotropic coordinate system defined by principal components of the covariance matrix $\C^t$. To update the evolution path $\vc{p}_{\sigma}^{t+1}$ a decay/relaxation factor $c_{\sigma}$ is used to decrease the importance of the previously performed steps with time.
	The step-size update rule increases the step-size if the length of the evolution path $\vc{p}_{\sigma}^{t+1}$ is longer than 
	the expected length of the evolution path under random selection $\mathbb{E} \left\| \NormOI \right\|$,
  and decreases otherwise (line \ref{CMAStepSizeUpdate}). The expectation of $\left\| \NormOI \right\|$ is approximated by $\sqrt{n} (1 - \frac{1}{4 n} + \frac{1}{21 n^2} )$.
	A damping parameter $d_{\sigma}$ controls the change of the step-size.
	
	The covariance matrix update consists of two parts (line \ref{CMAupdate}): \textit{rank-one update} \citep{2001HansenCMA} and \textit{ rank-$\mu$ update} \citep{2003HansenCMA}.	The rank-one update computes evolution path $\vc{p}_c^{t+1}$ of successful moves of the mean $\frac{\vc{m}^{t+1}-\vc{m}^{t}}{\sigma^{t}}$ 
	of the mutation distribution in the given coordinate system (line \ref{CMAEvoPathUpdate}), 
	in a similar way as for the evolution path $\vc{p}_{\sigma}^{t+1}$ of the step-size. 
	To stall the update of $\vc{p}_c^{t+1}$ when $\sigma$ increases rapidly, a $h_{\sigma}$ trigger is used (line \ref{CMAhsigma}).
	
	The rank-$\mu$ update computes a covariance matrix $\C_{\mu}$ as a weighted sum of covariances of successful steps of $\mu$ best individuals (line \ref{CMAplusCov}).
	The update of $\C$ itself is a replace of the previously accumulated information by a new one with corresponding weights of importance (line \ref{CMAupdate}):
	$c_1$ for covariance matrix $\vc{p}^{t+1}_c {\vc{p}^{t+1}_c}^T$ of rank-one update and 
	$c_{\mu}$ for $\C_{\mu}$ of rank-$\mu$ update \citep{2003HansenCMA} such that $c_1 + c_{\mu} \leq 1$.
	It was also proposed to take into account unsuccessful mutations in the \textit{"active" rank-$\mu$ update} \citep{2010HansenBBOBActiveCMA,2006ArnoldActiveCMA}.

  In CMA-ES, the factorization of the covariance $\C$ into $\A \A^T=\C$ is needed to sample the multivariate normal distribution (line \ref{CMAsampling}). The eigendecomposition with $O(n^3)$ complexity is used for the factorization. Already in the original CMA-ES it was proposed to perform the eigendecomposition every $n/10$ generations (not shown in \textbf{Algorithm} \ref{CMAdefault}) to reduce the complexity per function evaluation to $O(n^2)$. 
																														
\subsection{Large Scale Variants}\label{largescale}

The original $O(n^2)$ time and space complexity of CMA-ES precludes its applications to large scale optimization with $n\gg1000$. To enable the algorithm for large scale optimization, a linear time and space version called sep-CMA-ES was proposed by \cite{ros2008simple}. The algorithm does not learn dependencies but the scaling of variables by restraining the covariance matrix update to the diagonal elements: 

	\begin{eqnarray}
	c^{t+1}_{jj} = (1 - c_{cov}) c^t_{jj} + \frac{1}{\mu_{cov}}\left(\vc{p}_c^{t+1}\right)^2_j +
	 c_{ccov}\left(1-\frac{1}{\mu_{ccov}}\right) \sum_{i=1}^{\mu} w_i c^t_{jj} \left({z_{i:\lambda}}^{t+1}\right)^2_j, j=1,\ldots,n
	\end{eqnarray}
 where, for $j=1,\ldots,n$ the $c_{jj}$ are the diagonal elements of $\C^t$ and the $\left({z_{i:\lambda}}^{t+1}\right)_j = \left({x_{i:\lambda}}^{t+1}\right)_j/(\sigma^t \sqrt{c^t_{jj}})$.

This update reduces the computational complexity to $O(n)$ and allows to exploit problem separability. The algorithm demonstrated good performance on separable problems and even outperformed CMA-ES on non-separable Rosenbrock function for $n>100$.

A Natural Evolution Strategy (NES) variant, the Rank-One NES (R1-NES) by \cite{sun2011linear}, 
adapts the search distribution according to the natural gradient with a particular low rank parametrization of the covariance matrix,

	\begin{equation}
  \C = \sigma^2 (\I + \vc{u}\vc{u}^T),
  \end{equation}

where $u$ and $\sigma$ are the parameters to be adjusted. The adaptation of the predominant eigendirection $\vc{u}$ allows the algorithm to solve highly non-separable problems while maintaining only $O(n)$ time and $O(\mu n)$ space complexity. 
The use of the natural gradient in the derivation of the NES algorithm motivated 
a research which led to the formulation of the Information Geometric Optimization (IGO) framework by \cite{2011OllivierIGO}. 

The IGO framework was used to derive a similar to R1-NES algorithm called VD-CMA \citep{akimoto2014comparison} 
with the sampling distribution parametrized by a \newpage Gaussian model with the covariance matrix restricted as follows:

	\begin{equation}
	\label{VDCMAeq}
  \C = D (\I + \vc{u}\vc{u}^T)D,
  \end{equation}
	
	where $D$ is a diagonal matrix of dimension $n$ and $\vc{u}$ is a vector in $\R^n$. This model is able to represent a scaling for each
variable by $D$ and a principal component, which is generally not parallel to an axis, by $Dv$ \citep{akimoto2014comparison}. 
It has $O(n)$ time and $O(\mu n)$ space complexity but i) typically demonstrates a better performance than sep-CMA-ES and R1-NES and 
ii) can solve a larger class of functions \citep{akimoto2014comparison}.

A version of CMA-ES with a limited memory storage also called limited memory CMA-ES (L-CMA-ES) was proposed by \cite{knight2007reducing}. The L-CMA-ES uses the $m$ eigenvectors and eigenvalues spanning the $m$-dimensional
dominant subspace of the $n \times n$-dimensional covariance matrix \C. 
 The authors adapted a singular value decomposition updating 
 algorithm developed by \cite{brand2006fast} that allowed to avoid 
the explicit computation and storage of the covariance matrix. 
For $m < n$ the performance in terms of the number of function evaluations gradually decreases while enabling the search in $\R^n$ for $n>10,000$. However, the computational complexity of $O(m^2n)$ practically (for $m$ in order of $\sqrt{n}$ as suggested by \cite{knight2007reducing}) leads to the same limitations of $O(n^2)$ time complexity as in the original CMA-ES.

The ($\mu/\mu_{w},\lambda$)-Cholesky-CMA-ES proposed by \cite{2009Suttorp11CMA} is of special interest in this paper because the LM-CMA is based on this algorithm. The Cholesky-CMA represents a version of CMA-ES with rank-one update where instead of performing the factorization of the covariance matrix $\C^t$ into $\A^t {\A^t}^T=\C^t$, the Cholesky factor $\A^t$ and its inverse ${\A^t}^{-1}$ are iteratively updated. From \textbf{Theorem 1} \citep{2009Suttorp11CMA} it follows that if $\C^t$ is updated as 

	\begin{equation}
  \C^{t+1} = \alpha \C^t + \beta \vc{v}^t {\vc{v}^t}^T,
  \end{equation}
	
	where $\vc{v} \in \R^n$ is given in the decomposition form $\vc{v}^t = \A^t \vc{z}^t$, and $\alpha,\beta \in \R^+$, then for $\vc{z} \neq \vc{0}$ a Cholesky factor of the matrix $\C^{t+1}$ can be computed by
	
	\begin{equation} 
	\label{Aupdate}
  \A^{t+1} = \sqrt{\alpha} \A^t + \frac{\sqrt{\alpha}}{{\left\| \vc{z}^t \right\|}^2} \left( \sqrt{1 + \frac{\beta}{\alpha} {{\left\| \vc{z}^t \right\|}^2}} -1 \right) [\A^t \vc{z}^t] {\vc{z}^t}^T,
  \end{equation}
	
	for $\vc{z}_t=\vc{0}$ we have $\A^{t+1}=\sqrt{\alpha}\A^t$. From the \textbf{Theorem 2} \citep{2009Suttorp11CMA} it follows that if ${\A^{-1}}^{t}$ is the inverse of $\A^t$, then the inverse of $\A^{t+1}$ can be computed by

	\begin{equation}
	\label{Ainvupdate}
  {\A^{-1}}^{t+1} = \frac{1}{\sqrt{\alpha}} {\A^{-1}}^{t} - \frac{1}{\sqrt{\alpha}{\left\| \vc{z}^t \right\|}^2} \left( 1- \frac{1}{\sqrt{1 + \frac{\beta}{\alpha} {{\left\| \vc{z}^t \right\|}^2}}} \right) \vc{z}^t [{\vc{z}^t}^T {\A^{-1}}^{t}],
  \end{equation}
	
	for $\vc{z}^t \neq \vc{0}$ and by ${\A^{-1}}^{t+1}=\frac{1}{\sqrt{\alpha}} {\A^{-1}}^{t}$ for $\vc{z}^t = \vc{0}$.

\begin{algorithm}[tb!]
\caption{The ($\mu/\mu_{w},\lambda$)-Cholesky-CMA-ES}
\label{CholCMA}
\begin{algorithmic}[1]
\STATE{\textbf{given} $n \in \mathbb{N}_+$, $\lambda = 4 + \lfloor 3 \ln \, n  \rfloor $, $\mu =  \lfloor \lambda/2   \rfloor $, 
											$w_i = \frac{ \ln(\mu + 1) - \ln(i)}{\mu \ln(\mu + 1) - \sum_{j=1}^{\mu} \ln(j)}; i=1 \ldots \mu$,
											$\mu_w = \frac{1}{\sum^{\mu}_{i=1} w^2_i}$,
											$c_{\sigma} = \frac{\sqrt{\mu_{w}}}{
											\sqrt{n} + \sqrt{\mu_{w}}}$,      
											$d_{\sigma} = 1 + c_{\sigma} +2 \, \mstr{max}(0,\sqrt{ \frac{\mu_w - 1}{n+1}}-1)$,
											$c_c = \frac{4}{n+4}$, $c_1 = \frac{2}{ {(n + \sqrt{2})}^2}$
											} \label{CholCMAEScmaGiven}
\STATE{\textbf{initialize} $\vc{m}^{t=0} \in \R^{\dd}, \sigma^{t=0} > 0, \vc{p}^{t=-1}_{\sigma} = \ma{0}, \vc{p}^{t=-1}_{c} = \ma{0}, \A^{t=0} = \Id, \A^{t=0}_{inv} = \Id, t \leftarrow 0 $}
\REPEAT
  \FOR{$k \leftarrow 1,\ldots,\lambda$} \label{CholCMAGenerateBegin}
		\STATE{ $\vc{z}_k \leftarrow {{\mathcal N}  \hspace{-0.13em}\left({\ma{0},\I}\right)} $} \label{CholCMAsampling1}
			\STATE{ $\vc{x}_k \leftarrow \vc{m}^t + \sigma^{t} \A \vc{z}_k $} \label{CholCMAsampling2}
			\STATE{ $\vc{f}_k \leftarrow f(\vc{x}_k)$} \label{CholCMAGenerateEnd}
  \ENDFOR
	\STATE{ $ \vc{m}^{t+1} \leftarrow \sum_{i=1}^{\mu} \vc{w}_i \vc{x}_{i:\lambda} \;$}  \label{CholCMAComputeNewMean}
		\STATE{ $ \vc{z}_w \leftarrow \sum_{i=1}^{\mu} \vc{w}_i \vc{z}_{i:\lambda} \;$} \label{CholCMAComputeNewZMean}
	\STATE{ $ \vc{p}^{t}_{\sigma} \leftarrow (1 - c_{\sigma}) \vc{p}^{t-1}_{\sigma} + \sqrt{c_{\sigma}(2-c_{\sigma})} \sqrt{\mu_w} \vc{z}_w$} \label{CholCMASigmaPathUpdate}
	\STATE{ $ \vc{p}^{t}_{c} \leftarrow (1 - c_{c}) \vc{p}^{t-1}_{c} + \sqrt{c_{c}(2-c_{c})} \sqrt{\mu_w} \A^t \vc{z}_w$} \label{CholCMAEvoPathUpdate}
	\STATE{$ \vc{v}^t \leftarrow \A_{inv}^t \vc{p}^t_c  $}
	\STATE{$ \A^{t+1} \leftarrow \sqrt{1 - c_1} \A^t + \frac{\sqrt{1 - c_1}}{{\left\| \vc{v}^t \right\|}^2} \left( \sqrt{1 + \frac{c_1}{1 - c_1} {{\left\| \vc{v}^t \right\|}^2}} -1 \right) \vc{p}^t_c {\vc{v}^t}^T $} \label{CholAUpdate}
	\STATE{$ {\A_{inv}^{t+1}} \leftarrow \frac{1}{\sqrt{1 - c_1}} {\A_{inv}^t} - 
	\frac{1}{\sqrt{1 - c_1}{\left\| \vc{v}^t \right\|}^2} 
	\left( 1- \frac{1}{\sqrt{1 + \frac{c_1}{1 - c_1} {{\left\| \vc{v}^t \right\|}^2}}} \right) 
	{\vc{v}}^t [{{\vc{v}}^t}^T {\A_{inv}^t}], 
	$}						\label{CholAInvUpdate}
	\STATE{ $ \sigma^{t+1} \leftarrow \sigma^{t} \mstr{exp}
	          ( \frac{c_{\sigma}}{d_{\sigma}} (  \frac{\left\| \vc{p}^{t}_{\sigma} \right\|}{ \mathbb{E} \left\| \NormOI \right\| } - 1  )) $} \label{CholCMAStepSizeUpdate}
  \STATE{ $ t \leftarrow t + 1$}
\UNTIL{ \textit{stopping criterion is met} }
\end{algorithmic}
\end{algorithm}

The ($\mu/\mu_{w},\lambda$)-Cholesky-CMA-ES is outlined in \textbf{Algorithm} \ref{CholCMA}. 
As well as in the original CMA-ES, Cholesky-CMA-ES proceeds by sampling $\lambda$ candidate solutions (lines \ref{CholCMAGenerateBegin} - \ref{CholCMAGenerateEnd})  and taking into account the most successful $\mu$ out of $\lambda$ solutions in the evolution path adaptation (lines \ref{CholCMASigmaPathUpdate} and 
\ref{CholCMAEvoPathUpdate}). However, the eigendecomposition procedure is not required anymore because the Cholesky factor and its inverse are updated incrementally (line \ref{CholAUpdate} and \ref{CholAInvUpdate}). This simplifies a lot the implementation of the algorithm and keeps its time complexity as $O(n^2)$. A postponed update of the Cholesky factors every $O(n)$ iterations would not reduce the asymptotic  complexity further (as it does in the original CMA-ES) because the quadratic complexity will remain due to matrix-vector multiplications needed to sample new individuals. 

The non-elitist Cholesky-CMA is a good alternative to the original CMA-ES and demonstrates a comparable performance \citep{2009Suttorp11CMA}. While it has the same computational and memory complexity, the lack of rank-$\mu$ update may deteriorate its performance on problems where it is essential. 

\section{The LM-CMA}
\label{lmcmasection}

The LM-CMA is inspired by the L-BFGS algorithm but instead of storing $m$ gradients for performing inverse Hessian requiring operations
it stores $m$ direction vectors to reproduce the Cholesky factor $\A$ 
and generate candidate solutions with a limited time and space cost $O(mn)$ (see Section \ref{cholfactsection}).
 These $m$ vectors are estimates of descent directions provided by evolution path vectors and 
should be stored with a particular temporal distance to enrich $\A$ (see Section \ref{vecselectionsection}). 
 An important novelty introduced w.r.t. the original LM-CMA proposed by \cite{2014LoshchilovLMCMA} is 
a procedure for sampling from a family of search representations defined by Cholesky factors reconstructed from $m^* \leq m$ vectors (see Section \ref{metoile}) 
and according to the Rademacher distribution (see Section \ref{rademachsec}). 
These novelties allow to simultaneously reduce the internal time complexity of the algorithm and improve its performance in terms of the number of function evaluations. 
Before describing the algorithm itself, we gradually introduce all the necessary components. 

\subsection{Reconstruction of Cholesky Factors}
\label{cholfactsection}

By setting $a=\sqrt{1-c_1}$, $b^t=\frac{\sqrt{1-c_1}}{{\left\| \vc{v}^t \right\|}^2} \left( \sqrt{1 + \frac{c_1}{1-c_1} {{\left\| \vc{v}^t \right\|}^2}} -1 \right)$ and considering the \textit{evolution path} $\vc{p}^t_c$ (a change of optimum estimate $\vc{m}$ smoothed over iterations, see line \ref{LMCMAComputeNewMean} of \textbf{Algorithm} \ref{LMCMA}) together with $\vc{v}^t=\A^{{-1}^t}{\vc{p}}^t_c$, one can rewrite Equation (\ref{Aupdate}) as 

	\begin{equation} 
	\label{AupdateRew}
  \A^{t+1} = a \A^t + b^t \vc{p}_c^t {\vc{v}^t}^T,
  \end{equation}

The product of a random vector $\vc{z}$ and the Cholesky factor $\A^t$ thus can be directly computed. At iteration $t=0$, $\A^0=\I$ and $\A^0 \vc{z}=\vc{z}$, the new updated Cholesky factor $\A^{1}=a \I + b^{0} \vc{p}_c^{0} {\vc{v}^{0}}^T$. At iteration $t=1$, $\A^{1} \vc{z} = (a \I + b^{0} \vc{p}_c^{0} {\vc{v}^{0}}^T) \vc{z} = a \vc{z} + b^0 \vc{p}_c^0 ({\vc{v}^0}^T \vc{z})$ and $\A^2=a (a \I + b^{0} \vc{p}_c^{0} {\vc{v}^{0}}^T) + b^1 \vc{p}_c^1 {\vc{v}^1}^T$. Thus, a very simple iterative procedure which scales as $O(mn)$ can be used to sample candidate solutions in $\RR^n$ according to the Cholesky factor $\A^t$ reconstructed from $m$ pairs of vectors $\vc{p}_c^t$ and $\vc{v}^t$.
	
\begin{algorithm}[tb!]
\caption{ Az(): Cholesky factor - vector update}
\label{CholVector}
\begin{algorithmic}[1]
\STATE{\textbf{given} $\vc{z} \in \R^n,  m \in \mathbb{Z}_+, \vc{j} \in \mathbb{Z}_+^m, \vc{i} \in \mathbb{Z}^{\left|\vc{i}\right|}_+, \vc{P} \in \R^{m \times n}, \vc{V} \in \R^{m \times n}, \vc{b} \in \R^m, a \in [0,1]$
											} \label{CholVectorGiven1}
\STATE{\textbf{initialize} $\vc{x}  \leftarrow \vc{z}$}
  \FOR{$t \leftarrow 1,\ldots,\left|\vc{i}\right|$} \label{CholVectorGenerateBegin}
		\STATE{ $k \leftarrow \vc{b}^{\vc{j}_{\vc{i}_t}} \vc{V}^{(\vc{j}_{\vc{i}_t},:)} \cdot \vc{z} $} \label{CholVector1} 
		\STATE{ $\vc{x} \leftarrow  a \vc{x} + k \vc{P}^{(\vc{j}_{\vc{i}_t},:)}$} \label{CholVector3}
  \ENDFOR
\STATE{ \textbf{return} $\vc{x}$ }
\end{algorithmic}
\end{algorithm}

\begin{algorithm}[tb!]
\caption{ Ainvz(): inverse Cholesky factor - vector update}
\label{InvCholVector}
\begin{algorithmic}[1]
\STATE{\textbf{given} $\vc{z} \in \R^n, m \in \mathbb{Z}_+, \vc{j} \in \mathbb{Z}_+^m, \vc{i} \in \mathbb{Z}^{\left|\vc{i}\right|}_+, \vc{d} \in \R^m, c \in [0,1]$
											} \label{InvCholVectorGiven}
\STATE{\textbf{initialize} $\vc{x}  \leftarrow \vc{z}$}
  \FOR{$t \leftarrow 1,\ldots,\left|\vc{i}\right|$} \label{InvCholVectorGenerateBegin2}
		\STATE{ $k \leftarrow \vc{d}^{\vc{j}_{\vc{i}_t}}  \vc{V}^{(\vc{j}_{\vc{i}_t},:)} \cdot \vc{x} $} \label{InvCholVector1} 
		\STATE{ $\vc{x} \leftarrow  c \vc{x} - k \vc{V}^{(\vc{j}_{\vc{i}_t},:)}$} \label{InvCholVector3}
  \ENDFOR
\STATE{ \textbf{return} $\vc{x}$ }
\end{algorithmic}
\end{algorithm}		

The pseudo-code of the procedure to reconstruct $\vc{x}=\A^t \vc{z}$ from $m$ direction vectors 
is given in \textbf{Algorithm} \ref{CholVector}. At each iteration of reconstruction of $\vc{x}=\A^t \vc{z}$  (lines \ref{CholVectorGenerateBegin} - \ref{CholVector1}), $\vc{x}$ is updated as a sum of $a$-weighted version of itself and ${b}^t$-weighted evolution path $\vc{p}_c^t$ (accessed from a matrix $\vc{P} \in R^{m \times n}$ as $\vc{P}^{(\vc{i}_t,:)}$ ) scaled by the dot product of $\vc{v}^t$ and $\vc{x}$. As can be seen, \textbf{Algorithms} \ref{CholVector} and \ref{InvCholVector} use $\vc{j}_{\vc{i}_t}$ indexation instead of $t$. This is a convenient way to have references to matrices $\vc{P}$ and $\vc{V}$ which store $\vc{p}_c$ and $\vc{v}$ vectors, respectively. In the next subsections, we will show how to efficiently manipulate these vectors.
	
	A very similar approach can be used to reconstruct $\vc{x}={\A^t}^{-1} \vc{z}$, for the sake of reproducibility the pseudo-code is given in \textbf{Algorithm} \ref{InvCholVector} for $c=1/\sqrt{1-c_1}$ and $d^t= \frac{1}{\sqrt{1-c_1}{\left\| \vc{v}^t \right\|}^2} \times \left( 1- \frac{1}{\sqrt{1 + \frac{c_1}{1-c_1} {{\left\| \vc{v}^t \right\|}^2}}} \right)$. The computational complexity of both procedures scales as $O(mn)$.
	
	It is important to note that when a vector $\vc{p}^{\ell}$ from a set of $m$ vectors stored in $\vc{P}$ is replaced by a new vector $\vc{p}^{t+1}$ (see line \ref{updatesetline} in \textbf{Algorithm} \ref{LMCMA}), 
	\textit{all} inverse vectors $\vc{v}^t$ for $t=\ell,\ldots,m$ should be iteratively recomputed \citep{Krause2014}. 
	This procedure corresponds to line \ref{updateinversesline} in \textbf{Algorithm} \ref{LMCMA}.
	
\subsection{Direction Vectors Selection and Storage}	
\label{vecselectionsection}

The choice to store only $m\ll n$ direction vectors $\vc{p}_c$ to obtain a comparable amount of useful information as stored in the covariance matrix of the original CMA-ES requires a careful procedure of selection and storage. A simple yet powerful procedure proposed by \cite{2014LoshchilovLMCMA} is to preserve a certain temporal distance in terms of number of iterations between the stored direction vectors. The procedure tends to store a more  unique information in contrast to the case if the latest $m$ evolution path vectors would be stored. The latter case is different from the storage of $m$ gradients as in L-BFGS since the evolution path is gradually updated at each iteration 
with a relatively small learning rate $c_c$ and from $\mu \ll n$ selected vectors. 

\begin{algorithm}[tb!]
\caption{ UpdateSet(): direction vectors update }
\label{Selection}
\begin{algorithmic}[1]
\STATE{\textbf{given} $m \in \R^+, \vc{j} \in \mathbb{Z}_+^m, \vc{l} \in \mathbb{Z}_+^m, t \in \mathbb{Z}_+, \vc{N}\in \mathbb{Z}^m_+, \vc{P} \in \R^{m \times n}, \vc{p}_c \in \R^n$, $T \in \mathbb{Z}_+$
											} \label{sel1}
	\STATE{$t \leftarrow \left\lfloor t / T\right\rfloor$}
	\IF{$ t \leq m$}
		\STATE{$ \vc{j}_t \leftarrow t$}	\label{sel2}
	\ELSE
		\STATE{$ i_{min} \leftarrow  1+argmin_i\left(\vc{l}_{\vc{j}_{i+1}}-\vc{l}_{\vc{j}_i} - \vc{N}_{i}\right),|1\leq i \leq (m-1)$}	\label{sel3}
		\IF{$ \vc{l}_{\vc{j}_{i_{min}}}-\vc{l}_{\vc{j}_{i_{min}-1}} - \vc{N}_{i} \geq 0$}	\label{sel4}
			\STATE{$ i_{min} \leftarrow 1$}	\label{sel5}
		\ENDIF{}
			\STATE{$ \vc{j}_{tmp} \leftarrow \vc{j}_{i_{min}}$}	\label{sel7}
			\FOR{$i \leftarrow i_{min},\ldots,m-1$} \label{sel8}
				\STATE{ $\vc{j}_i \leftarrow \vc{j}_{i+1}$} \label{sel9} 
			\ENDFOR
			\STATE{ $\vc{j}_m \leftarrow \vc{j}_{tmp}$} \label{sel10} 
	\ENDIF{}
	\STATE{$j_{cur} \leftarrow \vc{j}_{\texttt{min}(t + 1, m)}$}	\label{sel12}
	\STATE{$\vc{l}_{j_{cur}} \leftarrow t T$}	\label{sel13}
	\STATE{$\vc{P}^{(\vc{j}_{cur},:)} \leftarrow \vc{p}_c$} \label{setpc}
\STATE{ \textbf{return}: $\vc{j}_{cur}$, $\vc{j}$, $\vc{l}$  }	\label{sel34}
\end{algorithmic}
\end{algorithm}	

The selection procedure is outlined in \textbf{Algorithm} \ref{Selection} which outputs an array of pointers $\vc{j}$ such that $\vc{j}_1$ points out to a row in matrices $\vc{P}$ and $\vc{V}$ with the oldest saved vectors $\vc{p}_c$ and $\vc{v}$ which will be taken into account during the reconstruction procedure. The higher the index $i$ of $\vc{j}_i$ the more recent the corresponding direction vector is. The index $j_{cur}$ points out to the vector which will be replaced by the newest one in the same iteration when the procedure is called. The rule to choose a vector to be replaced is the following. Find a pair of consecutively saved vectors
 ($\vc{P}^{(\vc{j}_{i_{min}-1},:)},\vc{P}^{(\vc{j}_{i_{min}},:)})$ with the distance between them (in terms of indexes of iterations, stored in $\vc{l}$) closest to a target distance $N_{i}$ (line \ref{sel3}). If this distance is smaller than $N_{i}$ then the index $j_{i_{min}}$ will be swapped with last index 
of $\vc{j}$ (lines \ref{sel7}-\ref{sel10}) and the corresponding 
vector $\vc{P}^{(\vc{j}_{i_{min}},:)}$ will be replaced by the new vector 
$\vc{p}_c$ (line \ref{setpc}), otherwise the oldest vector among $m$ saved vectors will be removed (as a result of line \ref{sel5}).
 Thus, the procedure gradually replaces vectors in a way to keep them at the distance $N_i$ and with the overall time horizon for all vectors of at most $\sum_i^{m-1} N_i$ iterations. 
The procedure can be called periodically every $T \in \mathbb{Z}_+$ iterations of the algorithm. 
The values of $N_i$ are to be defined, e.g., as a function of problem dimension $n$ and direction vector index $i$. 
Here, however, we set $N_i$ to $n$ for all $i$, i.e., the target distance equals to the problem dimension. 

\subsection{Sampling from a Family of Search Representations}
\label{metoile}

At iteration $t$, a new $k$-th solution can be generated as 

\begin{equation} 
\label{newsol}
 \vc{x}_k \leftarrow \vc{m}^t + \sigma^{t} Az( \vc{z}_k, \vc{i} ),
\end{equation}

where $\vc{z}_k \in \R^n$ is a vector drawn from some distribution and 
transformed by a Cholesky factor by calling $Az(\vc{z}_k,\vc{i})$. 
The $Az()$ procedure (see \textbf{Algorithm} \ref{CholVector}) has an input $\vc{i}$
 which defines indexes of direction vectors used to reconstruct 
the Cholesky factor. 
It is important to note that $\vc{P}^{(1,:)}$ refers to the first vector physically stored in matrix $\vc{P}$,  $\vc{P}^{(\vc{j}_1,:)}$ refers to the oldest vector, $\vc{P}^{(\vc{j}_{\vc{i}_t},:)}$ refers to the $\vc{i}_t$-th oldest vector according to an array $\vc{i}$ with indexes of vectors of interest. 
Thus, by setting $\vc{i} = 1,\ldots,m$ all $m$ vectors will be used in the reconstruction. Importantly, omission of some vector in $\vc{i}$ can be viewed as 
setting of the corresponding learning rate in Equation (\ref{AupdateRew}) to 0.

By varying $\vc{i}$, one can control the reconstruction of the Cholesky factor used for sampling and in this way explore a family of possible transformations of the coordinate system. The maximum number of vectors defined by $m$ can be associated with the number of degrees of freedom of this exploration. 

While in the original LM-CMA \citep{2014LoshchilovLMCMA} the value of $m$ is set to $4 + \lfloor 3 \ln \, n  \rfloor $ to allow the algorithm scale up to millions of variables, we found that greater values of $m$, e.g., $\sqrt{n}$ often lead to better performance (see Section \ref{sectsscalingm} for a detailed analysis). 
Thus, when memory allows, a gain in performance can be achieved.
However, due to an internal cost $O(mn)$ of $Az()$, the time cost then would scale as $O(n^{3/2})$ which is undesirable for $n \gg 1000$. 
This is where the use of $m^{*}$ out of $m$ vectors can drastically reduce the time complexity. We propose to sample $m^{*}$ from a truncated half-normal distribution $\left| {{\mathcal N}  \hspace{-0.13em}\left({{0},m^2_{\sigma}}\right)} \right|$ (see line \ref{selectmetoile} of \textbf{Algorithm} \ref{SelectSub}) and set $\vc{i}$ to the latest $m^{*}$ vectors (line \ref{setivals}). For a constant $m_{\sigma}=4$, the time complexity of $Az()$ scales as $O(n)$. New value of $m^{*}$ is generated for each new individual. Additionally, to exploit the oldest information, we force $m^{*}$ to be generated with $10m_{\sigma}$ for one out of $\lambda$ individuals. 
While for $m^{*}=0$ the new solution $\vc{x}_k$ appears to be sampled 
from an isotropic normal distribution, the computation of $\vc{v}$ inverses is performed using all $m$ vectors. 

\begin{algorithm}[tb!]
\caption{ SelectSubset(): direction vectors selection}
\label{SelectSub}
\begin{algorithmic}[1]
\STATE{\textbf{given} $m \in \mathbb{Z}_+, m_{\sigma} = 4,k \in \mathbb{Z}_+$							}
	\IF{$k = 1$}
			\STATE{$ m_{\sigma} \leftarrow 10 m_{\sigma}$}
	\ENDIF
			\STATE{$ m^{*} \leftarrow \min(\left\lfloor m_{\sigma}  \left| {{\mathcal N}  \hspace{-0.13em}\left({{0},{1}}\right)} \right| \right\rfloor, m)$}	\label{selectmetoile}
			\STATE{$\vc{i} \leftarrow (m+1-m^{*}),\ldots,m$}	\label{setivals}
\STATE{ \textbf{return} $\vc{i}$ }
\end{algorithmic}
\end{algorithm}

\subsection{Sampling from the Rademacher Distribution}
\label{rademachsec}

Evolution Strategies are mainly associated with the multivariate normal distribution used to sample candidate solutions. 
However, alternative distributions such as the Cauchy distribution can be used \citep{yao1997fast,2008HansenAdaptiveEncoding}. 
Moreover, the Adaptive Encoding procedure proposed by \cite{2008HansenAdaptiveEncoding} can be coupled with any sampling distribution as in \cite{loshchilov2011adaptive}, where it was shown that completely deterministic adaptive coordinate descent on principal components obtained with the Adaptive Encoding procedure allows to obtain the performance comparable to the one of CMA-ES. 

In this paper, inspired by \cite{loshchilov2011adaptive}, we replace the original multivariate normal distribution used in LM-CMA by the Rademacher distribution, where a random variable has 50\% chance to be either -1 or +1 (also can be viewed as a Bernoulli distribution). Thus, a pre-image vector of candidate solution $\vc{z}$ contains $n$ values which are either -1 or +1. 
Our intention to use this distribution is three-fold: i) to reduce the computation complexity by a constant but rather significant factor, 
ii) to demonstrate that the Rademacher distribution can potentially be an alternative to the Gaussian distribution at least in large scale 
settings, 
iii) to demonstrate that our new step-size adaptation rule (see next section), which does not make assumptions about the sampling distribution, can work well when used with non-Gaussian distributions. 
As a support for this substitution, we recall that for a $n$-dimensional unit-variance spherical Gaussian, 
for any positive real number $\beta \leq \sqrt{n}$, all but at most $3\exp^{-c \beta^2}$ of the mass lies within the annulus 
$\sqrt{n-1}-\beta \leq r \leq \sqrt{n-1}+\beta$, where, $c$ is a fixed positive constant \citep{hopcroft2014foundations}. 
Thus, when $n$ is large, the mass is concentrated in a thin annulus of width $O(1)$ at radius $\sqrt{n}$. 
Interestingly, the sampling from the Rademacher distribution reproduces this effect of large-dimensional Gaussian sampling 
since the distance from the center of a $n$-dimensional hypercube to its corners is $\sqrt{n}$.

\subsection{Population Success Rule}	
\label{PSRsection}

The step-size used to define the scale of deviation of a sampled candidate solution from the mean of the mutation distribution can be adapted by the Population Success Rule (PSR) proposed for LM-CMA by \cite{2014LoshchilovLMCMA}. This procedure does not assume that candidate solutions should come from the multivariate normal distribution as it is often assumed in Evolution Strategies including CMA-ES. Therefore, PSR procedure is well suited for the Rademacher distribution. 

The PSR is inspired by the \textit{median success rule} \citep{elhara2013median}. To estimate the success of the current population we combine fitness function values from the previous and current population into a mixed set 
	
	\begin{equation} 
	\label{Fmix}
  \vc{f}_{mix} \leftarrow \vc{f}^{\,t-1} \cup \vc{f}^{\,t}
  \end{equation}
	
	Then, all individuals in the mixed set are ranked to define two sets $\vc{r}_{t-1}$ and $\vc{r}_{t}$ (the lower the rank the better the individual) containing ranks of individuals of the previous and current populations ranked in the mixed set. 	
	A normalized success measurement is computed as
	
	\begin{equation} 
	\label{ZPSR}
  z_{PSR} \leftarrow \frac{\sum_{i=1}^{\lambda} \vc{r}^{t-1}(i) - \vc{r}^{t}(i)}{\lambda^2} - z^{*},
  \end{equation}
	where $z^*$ is a target success ratio and $\lambda^2$ accounts for the normalization of the sum term 
	and for different possible population size $\lambda$. 
	Then, for $s \leftarrow (1 - c_{\sigma}) s + c_{\sigma} z_{PSR}$, the step-size is adapted as 
	
	\begin{equation} 
	\label{MruleSigma}
  \sigma^{t+1} \leftarrow \sigma^t \exp\left({s}\right / d_{\sigma}),
  \end{equation}
	
	where $d_{\sigma}$ is a damping factor which we set here to 1.

\begin{algorithm}[tb!]
\caption{The ($\mu/\mu_{w},\lambda$)-LM-CMA}
\label{LMCMA}
\begin{algorithmic}[1]
\STATE{\textbf{given} $n \in \mathbb{N}_+$, $\lambda = 4 + \lfloor 3 \ln \, n  \rfloor $, $\mu =  \lfloor \lambda/2   \rfloor $, 
											$w_i = \frac{ \ln(\mu + 1) - \ln(i)}{\mu \ln(\mu + 1) - \sum_{j=1}^{\mu} \ln(j)}; i=1 \ldots \mu$,
											$\mu_w = \frac{1}{\sum^{\mu}_{i=1} w^2_i}$,
											$c_{\sigma} = 0.3$, $z^{*} = 0.25$,      
											$m = 4 + \lfloor 3 \ln \, n  \rfloor $, $N_{steps} = n$,
											$c_c = \frac{0.5}{\sqrt{n}}$, $c_1 = \frac{1}{10\ln(n+1)}$, $d_{\sigma}=1$, 
											$T=\left\lfloor log(n)\right\rfloor$
											} \label{LMCMAGiven}
\STATE{\textbf{initialize} $\vc{m}^{t=0} \in \R^{\dd}, \sigma^{t=0} > 0,  \vc{p}^{t=0}_{c} = \ma{0}, s \leftarrow 0 , t \leftarrow 0 $}
\REPEAT
  \FOR{$k \leftarrow 1,\ldots,\lambda$} \label{LMCMAGenerateBegin}
		\IF{$k \pmod 2 = 1$}
			
			\STATE{$ \vc{z}_k \leftarrow Rademacher() $}
			\STATE{$ \vc{i} \leftarrow SelectSubset(k) $}
			\STATE{ $\vc{x}_k \leftarrow \vc{m}^t + \sigma^{t} Az( \vc{z}_k, \vc{i} ) $} \label{LMCMAsampling2}			
		\ELSE
			\STATE{$\vc{x}_k \leftarrow \vc{m}^t - (\vc{x}_{k-1} - \vc{m}^t) $}
		\ENDIF
		\STATE{ $\vc{f}^t_k \leftarrow f(\vc{x}_k)$} \label{LMCMAGenerateEnd}
  \ENDFOR
	\STATE{ $ \vc{m}^{t+1} \leftarrow \sum_{i=1}^{\mu} \vc{w}_i \vc{x}_{i:\lambda} \;$}  \label{LMCMAComputeNewMean}
	\STATE{ $ \vc{p}^{t+1}_{c} \leftarrow (1 - c_{c}) \vc{p}^{t}_{c} + \sqrt{c_{c}(2-c_{c})} \sqrt{\mu_w} (\vc{m}^{t+1} - \vc{m}^t) / {\sigma^t}$} \label{LMCMAEvoPathUpdate}
\IF{$t \pmod{T} = 0$} 
	\STATE{$ UpdateSet(\vc{p}^{t+1}_{c}) $} \label{updatesetline}
	\STATE{$ UpdateInverses() $}	\label{updateinversesline}
\ENDIF
	\STATE{ $ \vc{r}^{t}, \vc{r}^{t-1} \leftarrow$ Ranks of $\vc{f}^{\,t}$ and $\vc{f}^{\,t-1}$ in $\vc{f}^{\,t} \cup \vc{f}^{\,t-1}$} \label{SuccRule1}
	\STATE{$ z_{PSR} \leftarrow \frac{\sum_{i=1}^{\lambda} \vc{r}^{t-1}(i) - \vc{r}^{t}(i)}{\lambda^2} - z^{*} $}
	\STATE{ $ s \leftarrow (1 - c_{\sigma})s + c_{\sigma}z_{PSR} $}
	\STATE{ $ \sigma^{t+1} \leftarrow \sigma^{t} \mstr{exp}
	          ( s / d_{\sigma}) $} \label{LMCMAsigma}
  \STATE{ $ t \leftarrow t + 1$}
\UNTIL{ \textit{stopping criterion is met} }
\end{algorithmic}
\end{algorithm}	

\subsection{The Algorithm}

The improved LM-CMA is given in \textbf{Algorithm} \ref{LMCMA}. 
At each iteration $t$, $\lambda$ candidate solutions are generated by mutation defined as a product of a vector $\vc{z}_k$ sampled from the Rademacher distribution and a Cholesky factor $\A^t$ reconstructed from $m^{*}$ out of $m$ vectors (line \ref{LMCMAGenerateBegin}-\ref{LMCMAGenerateEnd}) as described in Sections \ref{cholfactsection}-\ref{rademachsec}. We introduce the mirrored sampling \citep{brockhoff2010mirrored} to generate the actual $\vc{x}_k$ only every second time and thus decrease the computation cost per function evaluation by a factor of two 
by evaluating $\vc{m}^t + \sigma^{t} Az( \vc{z}_k)$ and then its mirrored version 
$\vc{m}^t - (\vc{x}_{k-1} - \vc{m}^t)$. The latter approach sometimes also improves the convergence rate.

The best $\mu$ out of $\lambda$ solutions are selected to compute the new mean $\vc{m}^{t+1}$ of the mutation distribution in line \ref{LMCMAComputeNewMean}. 
The new evolution path $\vc{p}^{t+1}_{c}$ is updated (line \ref{LMCMAEvoPathUpdate}) from the change of the mean vector $\sqrt{\mu_w} (\vc{m}^{t+1} - \vc{m}^t) / {\sigma^t}$ and represents an estimate of descent direction. 
One vector among $m$ vectors is selected and replaced by the new $\vc{p}^{t+1}_{c}$ 
in \textit{UpdateSet()} procedure described in Section \ref{vecselectionsection}. All inverses $\vc{v}$ of evolution path vectors which are at least as recent as the direction vector to be replaced should be recomputed in the \textit{UpdateInverses()} procedure as described in Section \ref{cholfactsection}. The step-size is updated according to the PSR rule described in Section \ref{PSRsection}.

\newcommand{\sphere}{$f_{Sphere}$}
\newcommand{\rosen}{$f_{Rosen}$}
\newcommand{\cigar}{$f_{Cigar}$}
\newcommand{\discus}{$f_{Discus}$}
\newcommand{\elli}{$f_{Elli}$}
\newcommand{\diffpow}{$f_{DiffPow}$}
\newcommand{\rotrosen}{$f_{RotRosen}$}
\newcommand{\rotcigar}{$f_{RotCigar}$}
\newcommand{\rotdiscus}{$f_{RotDiscus}$}
\newcommand{\rotelli}{$f_{RotElli}$}
\newcommand{\rotdiffpow}{$f_{RotDiffPow}$}

 \begin{table}[tb!]
\caption{Test functions, initialization intervals and initial standard deviation (when applied). $\vc{R}$ is an orthogonal $n \times n$ matrix with each column vector $\vc{q}_i$ being a uniformly distributed unit vector implementing an angle-preserving transformation \citep{ros2008simple}}.
\centering
\scriptsize
\begin{tabular}{lllll  cll  cr}
 \hline  
 Name 	& Function & Target $f(\vc{x})$ &Init & $\sigma^{0}$ \\ 
\hline
Sphere & \sphere(\vc{x})$=\sum_{i=1}^n \vc{x}^{2}_{i}$ & $10^{-10}$ & $[-5,5]^{n}$ & 3 \\ [0.10cm]
Ellipsoid & \elli(\vc{x})$=\sum_{i=1}^n 10^{6{\frac{i-1}{n-1}}} \vc{x}^{2}_{i}$ & $10^{-10}$ & $[-5,5]^{n}$ & 3 \\ [0.10cm]
Rosenbrock & \rosen(\vc{x})$=\sum_{i=1}^{n-1}\left(100.(\vc{x}^{2}_{i}-x_{i+1})^{2}+(x_{i}-1)^{2}\right)$ & $10^{-10}$ & $[-5,5]^{n}$ & 3 \\ [0.10cm]
Discus & \discus(\vc{x})$=10^6\vc{x}^2_1 + \sum_{i=2}^n \vc{x}^{2}_{i}$ & $10^{-10}$ & $[-5,5]^{n}$ & 3 \\ [0.10cm]
Cigar & \cigar(\vc{x})$=\vc{x}^2_1 + 10^6\sum_{i=2}^n \vc{x}^{2}_{i}$ & $10^{-10}$ & $[-5,5]^{n}$ & 3 \\ [0.10cm]
Different Powers & \diffpow(\vc{x})$=\sum_{i=1}^n \left|\vc{x}_i\right|^{2+4(i-1)/(n-1)}$ & $10^{-10}$ & $[-5,5]^{n}$ & 3 \\ [0.10cm]
Rotated Ellipsoid & \rotelli(\vc{x})=\elli(\vc{R}\vc{x}) & $10^{-10}$ & $[-5,5]^{n}$ & 3 \\ [0.10cm]
Rotated Rosenbrock & \rotrosen(\vc{x})=\rosen(\vc{R}\vc{x}) & $10^{-10}$ & $[-5,5]^{n}$ & 3 \\ [0.10cm]
Rotated Discus & \rotdiscus(\vc{x})=\discus(\vc{R}\vc{x}) & $10^{-10}$ & $[-5,5]^{n}$ & 3 \\ [0.10cm]
Rotated Cigar & \rotcigar(\vc{x})=\cigar(\vc{R}\vc{x}) & $10^{-10}$ & $[-5,5]^{n}$ & 3 \\ [0.10cm]
Rotated Different Powers & \rotdiffpow(\vc{x})=\diffpow(\vc{R}\vc{x}) & $10^{-10}$ & $[-5,5]^{n}$ & 3 \\ [0.10cm]
\hline
\end{tabular}
\label{table:TableFunc}
\end{table}

\section{Experimental Validation}
\label{resultssection}

The performance of the LM-CMA is investigated comparatively to the L-BFGS \citep{liu1989limited}, the active CMA-ES by \cite{2010HansenBBOBActiveCMA} and the VD-CMA by \cite{akimoto2014comparison}.  The sep-CMA-ES is removed from the comparison due to its similar but inferior performance w.r.t. the VD-CMA observed both in our study and by \cite{akimoto2014comparison}. 

We use the L-BFGS implemented in MinFunc library by \cite{schmidt2005minfunc} in its default parameter settings \footnote{\url{http://www.cs.ubc.ca/~schmidtm/Software/minFunc.html}}, 
the active CMA-ES (aCMA) without restarts in its default parametrization of CMA-ES MATLAB code version 3.61 \footnote{\url{http://www.lri.fr/~hansen/cmaes.m}}. 
For faster performance in terms of CPU time, the VD-CMA was (exactly) reimplemented in C language from the MATLAB code provided by the authors. 
For the sake of reproducibility, the source code of all algorithms is available online \footnote{\url{http://sites.google.com/site/ecjlmcma/}}. The default parameters of LM-CMA are given in \textbf{Algorithm} \ref{LMCMA}.

We use a set of benchmark problems (see Table \ref{table:TableFunc}) commonly used in Evolutionary Computation, more specifically in the BBOB framework \citep{2009FinckBBOB2009setup}. Indeed, many problems are missing including the ones where tested methods and LM-CMA fail to timely demonstrate reasonable performance in large scale settings. We focus on algorithm performance w.r.t. both the number of function evaluations used to reach target values of $f$, CPU time spent per function evaluation and the number of memory slots required to run algorithms. Any subset of these metrics can dominate search cost in large scale settings, while in low scale settings memory is typically of a lesser importance.

In this section, we first investigate the scalability of the proposed algorithm w.r.t. the existing alternatives.  
While both the computational time and space complexities scale moderately with problem dimension, the algorithm is capable to preserve certain 
invariance properties of the original CMA-ES. Moreover, we obtain unexpectedly good results on some well-known  benchmark problems, 
e.g., linear scaling of the budget of function evaluations to solve Separable and Rotated Ellipsoid problems. 
We demonstrate that the performance of LM-CMA is comparable to the one of L-BFGS with exact estimation of gradient information. 
Importantly, we show that LM-CMA is competitive to L-BFGS in very large scale (for derivative-free optimization) settings with 100,000 variables. 
Finally, we investigate the sensitivity of LM-CMA to some key internal parameters such as the number of stored direction vectors $m$. 

\subsection{Space Complexity}

\begin{figure}[t]
\centerline{ 
	\includegraphics[width=0.85\textwidth]{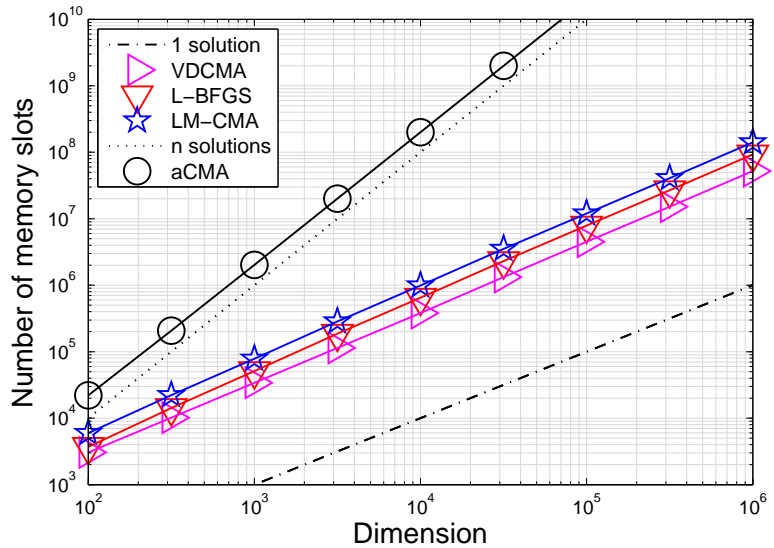} 
}
\caption{\label{fig:memory} Number of memory slots of floating point variables required to run different optimization algorithms versus problem dimension $n$.
}
\end{figure}

The number of memory slots required to run optimization algorithms versus problem dimension $n$ referred to as space complexity can limit applicability of certain algorithms to large scale optimization problems. Here, we list the number of slots up to constant and asymptotically constant terms.

The presented algorithms store $\lambda$ generated solutions (LM-CMA, VD-CMA and aCMA with the default $\lambda=4 + \lfloor 3 \ln \, n  \rfloor$ and $\mu=\lambda/2$, L-BFGS with $\lambda=1$) and some intermediate information (LM-CMA and L-BFGS with $m$ pairs of vectors, aCMA with at least two matrices of size $n\times n$) to perform the search. 
Our implementation of VD-CMA requires $(max(\mu+15,\lambda) + 7)n$ slots compared to $(2.5\lambda +21) n$ of the original MATLAB code. The LM-CMA requires $(2m + \lambda + 6)n + 5m$ slots, the L-BFGS requires $(2m + 3) n$ slots and aCMA requires $(2n + \lambda + 3)n$ slots. 

Figure \ref{fig:memory} shows that due to its quadratic space complexity aCMA 
requires about $2\times 10^8$ slots (respectively, $8\times 10^8$ slots) for 10,000-dimensional (respectively, 20,000-dimensional) problems which with 8 bytes per double-precision floating point number would correspond to about 1.6 GB (respectively, 6.4 GB) of computer memory. This simply precludes the use of CMA-ES and its variants with explicit storage of the full covariance matrix or Cholesky factors to large scale optimization problems with $n>10,000$. 
LM-CMA stores $m$ pairs of vectors as well as the L-BFGS. For $m=4 + \lfloor 3 \ln \, n  \rfloor$ (as the default population size in CMA-ES), L-BFGS is 2 times and LM-CMA is 3 times more expensive in memory than VD-CMA, but they all basically can be run for millions of variables. 

In this paper, we argue that additional memory can be used while it is allowed and is at no cost.
Thus, while the default $m=4 + \lfloor 3 \ln \, n  \rfloor$, we suggest to use $m=\lfloor2\sqrt{n}\rfloor$ if memory allows (see Section \ref{sectsscalingm}). In general, the user can provide a threshold on memory and if, e.g., by using $m=\lfloor2\sqrt{n}\rfloor$ this memory threshold would be violated, the algorithm automatically reduces $m$ to a feasible $m_f$.

\begin{figure}[t]
\centerline{ 
	\includegraphics[width=0.85\textwidth]{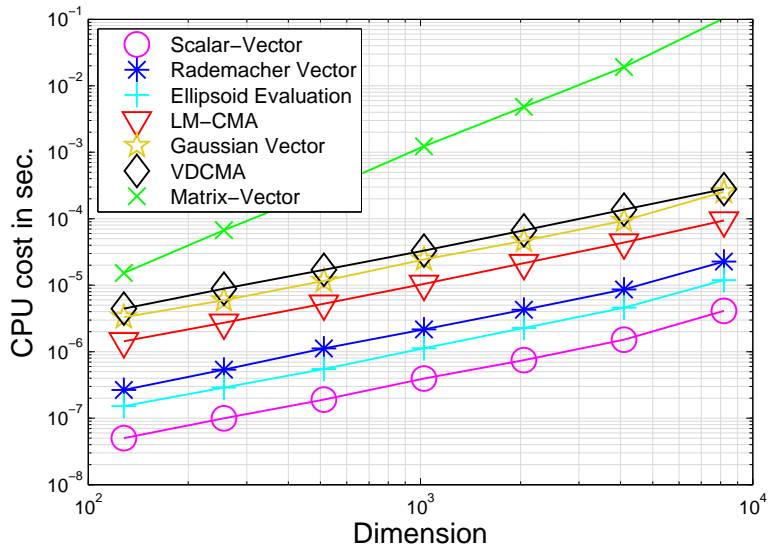} 
}
\caption{\label{fig:timing} Timing results of LM-CMA and VD-CMA averaged over the whole run on the separable Ellipsoid compared to timing results of simple operations averaged over 100 seconds of experiments. The results for L-BFGS are not shown but for an optimized implementation would be comparable to one scalar-vector multiplication.
}
\end{figure}

\subsection{Time Complexity}
\label{timespace}

The average amount of CPU time internally required by an algorithm per evaluation of some objective function $f \in \R^n$ (not per algorithm iteration) referred to as time complexity also can limit applicability of certain algorithms to large scale optimization problems. They simply can be too expensive to run, e.g., much more expensive than to perform function evaluations.

Figure \ref{fig:timing} shows how fast CPU time per evaluation scales for different operations measured on one 1.8 GHz processor of an Intel Core i7-4500U. Scalar-vector multiplication of a vector with $n$ variables scales linearly with ca. $4\cdot{10}^{-10}n$ seconds, evaluation of the separable Ellipsoid function is 2-3 times more expensive if a temporary data is used.
 Sampling of $n$ normally distributed variables scales as ca. 60 vectors-scalar multiplications that defines the cost of sampling of unique candidate solutions of many Evolution Strategies such as separable CMA-ES and VD-CMA. However, the sampling of variables according to the Rademacher distribution is about 10 times cheaper. The use of mirrored sampling also decreases the computational burden without worsening the convergence. Finally, the internal computation cost of LM-CMA scales linearly as about 25 scalar-vector multiplications per function evaluation. It is much faster than the lower bound for the original CMA-ES defined by one matrix-vector multiplication required to sample one individual. We observe that the cost of one matrix-vector multiplications costs about $2n$ scalar-vector multiplications, the overhead is probably due to access to matrix members. 

The proposed version of LM-CMA is about 10 times faster internally than the original version by \cite{2014LoshchilovLMCMA} due to the use of mirrored sampling, the Rademacher sampling distribution and sampling with $m^*$ instead of $m$ direction vectors both for $m=4 + \lfloor 3 \ln \, n  \rfloor$ and $m=\lfloor 2\sqrt{n} \rfloor$. For 8192-dimensional problems it is about 1000 times faster internally than CMA-ES algorithms with the full covariance matrix update (the cost of Cholesky-CMA-ES is given in \cite{2014LoshchilovLMCMA}).

\subsection{Invariance under Rank-preserving Transformations of the Objective Function}
\label{invrank}

The LM-CMA belongs to a family of so-called comparison-based algorithms. 
The performance of these algorithms is unaffected by rank-preserving (strictly monotonically increasing) transformations of the objective function, e.g., 
whether the function $f$, $f^3$ or $f \times \left| f \right|^{-2/3}$ is minimized \citep{2011OllivierIGO}. 
Moreover, this invariance property provides robustness to noise as far as this noise 
does not impact a comparison of solutions of interest \citep{auger2013linear}. 

In contrast, gradient-based algorithms are sensitive to rank-preserving transformations of $f$. While the availability of gradient information may 
mitigate the problem that objective functions with the same contours can be solved with a different number of functions evaluations, the lack of gradient information 
forces the user to estimate it with approaches whose numerical stability is subject to scaling of $f$. Here, we simulate L-BFGS in an idealistic black-box scenario when gradient information is estimated perfectly (we provide exact gradients) but at the cost of $n+1$ function evaluations per gradient that corresponds to the cost of the forward difference method. Additionally, we investigate the performance of L-BFGS with the central difference method ($2n+1$ evaluations per gradient) which is twice more expensive but numerically more stable. We denote this method as CL-BFGS.

\subsection{Invariance under Search Space Transformations}
\label{invdec}

\begin{figure}[!ht]%
	\includegraphics[width=0.5\textwidth]{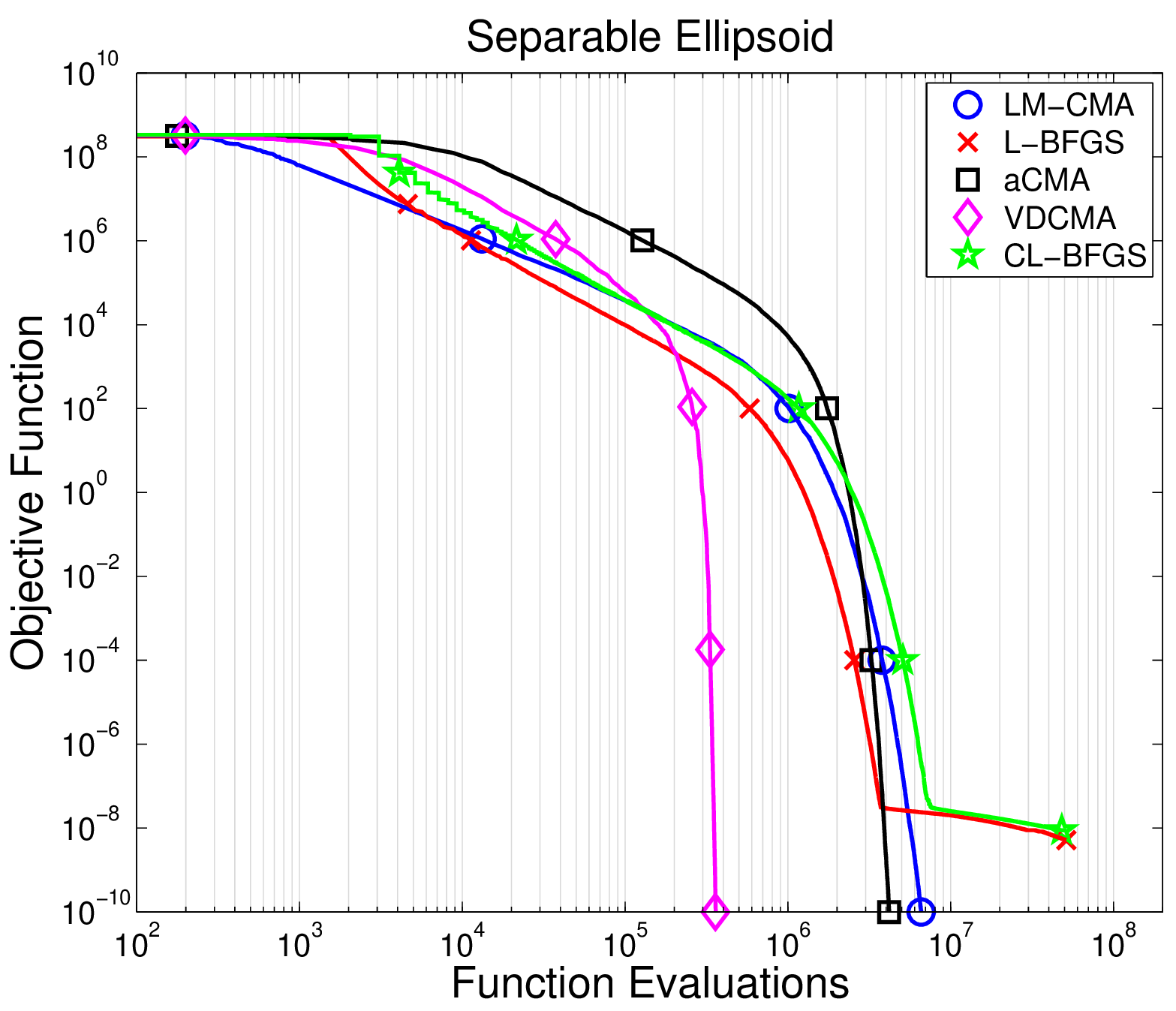}
  \includegraphics[width=0.5\textwidth]{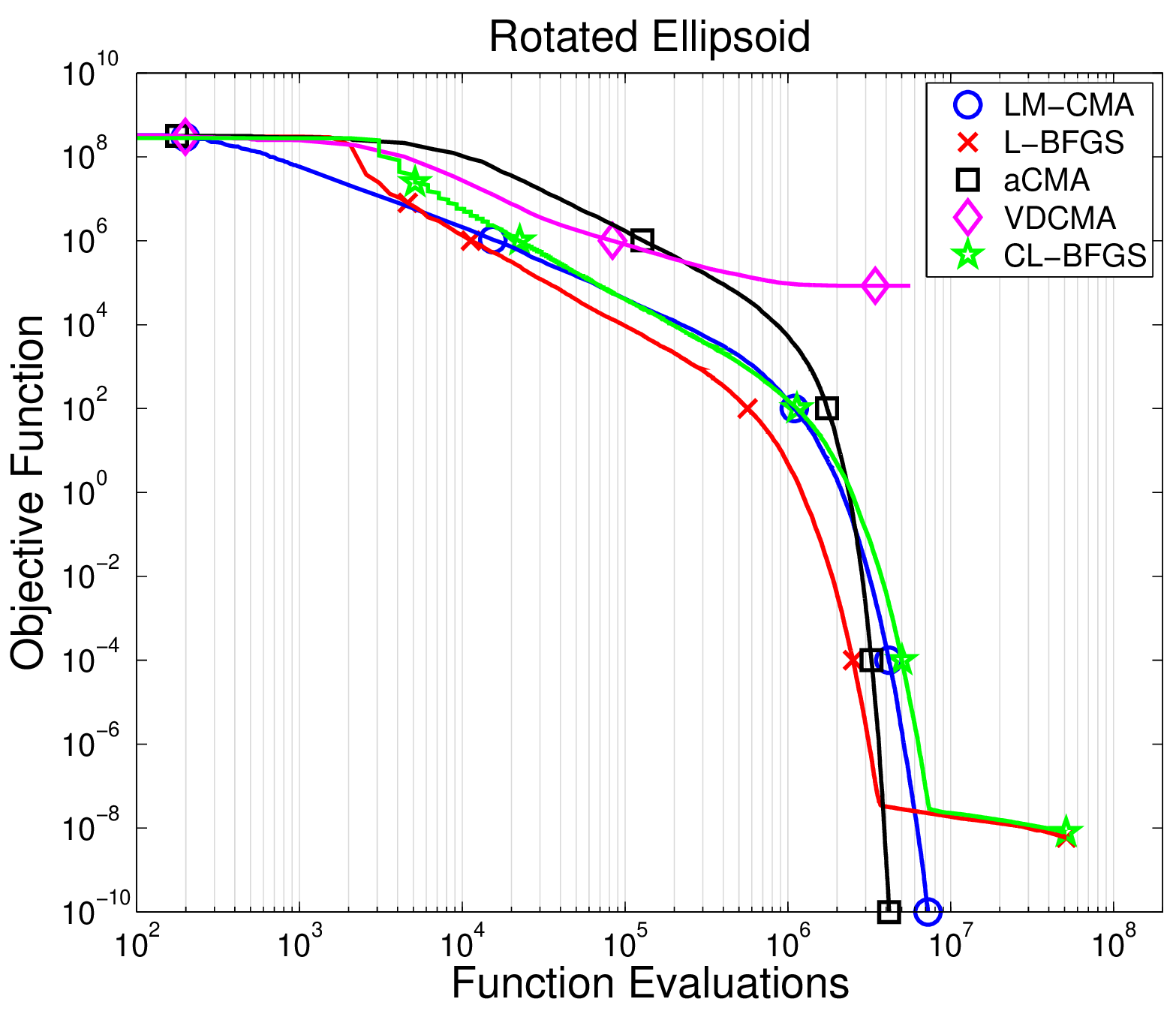}\\ 
	\includegraphics[width=0.5\textwidth]{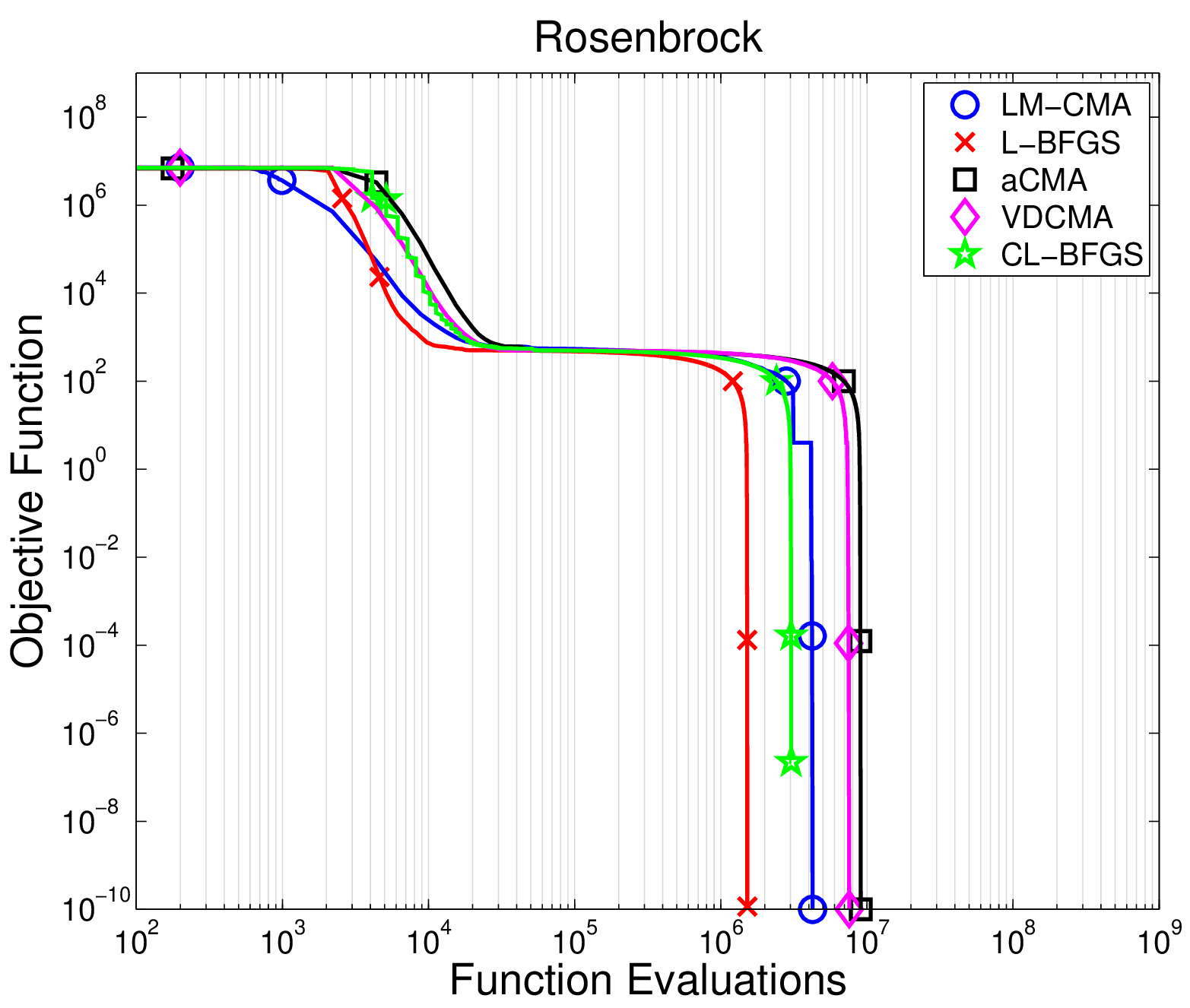}
  \includegraphics[width=0.5\textwidth]{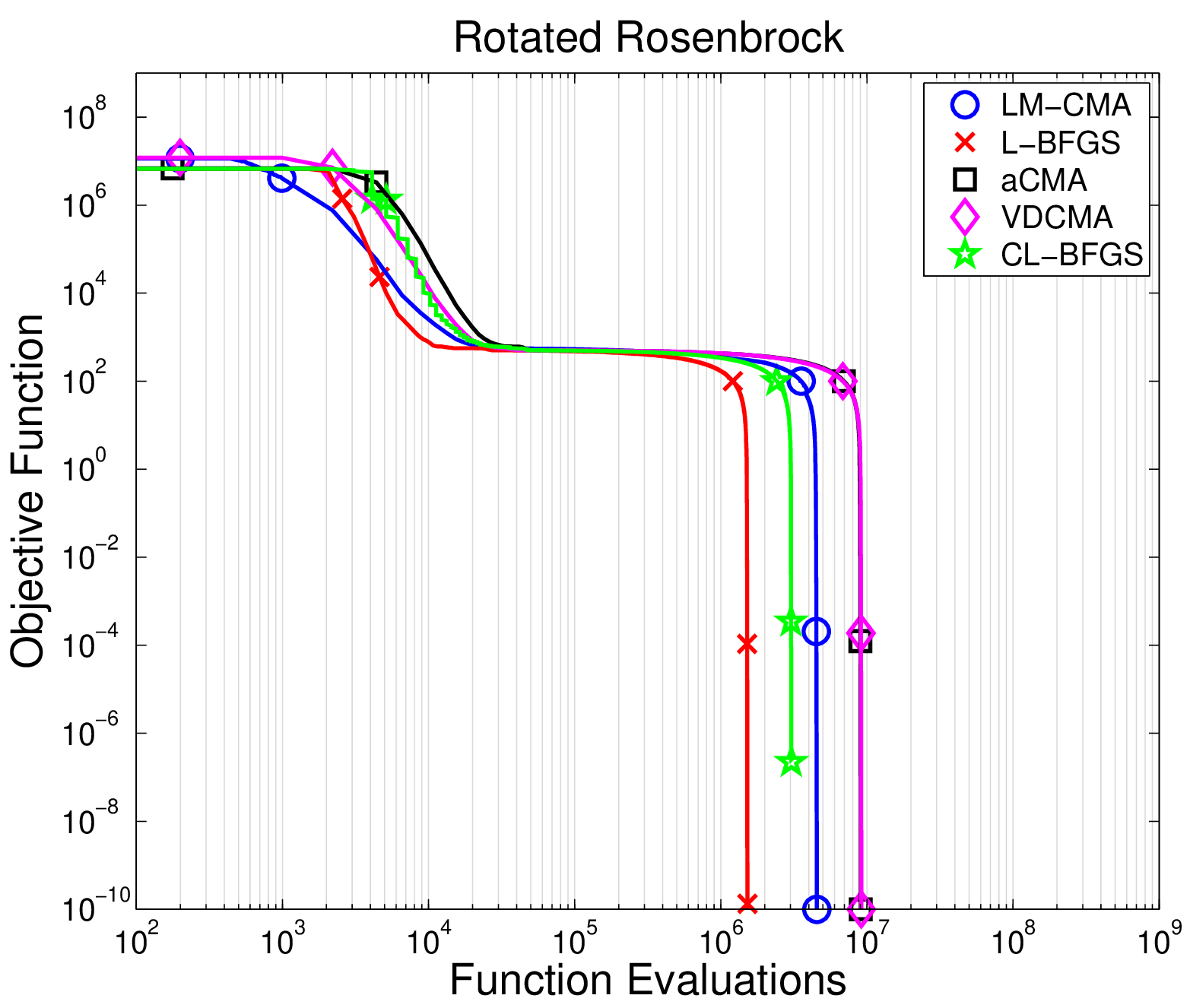}
\caption{\label{fig1} The trajectories show the median of 11 runs of LM-CMA, L-BFGS with exact gradients provided at the cost of $n+1$ evaluations per gradient, CL-BFGS with central differencing, active CMA-ES and VD-CMA  on 512- Separable/Original (Left Column) and Rotated (Right Column) functions.}
\end{figure}

\begin{figure}[!ht]
	\includegraphics[width=0.5\textwidth]{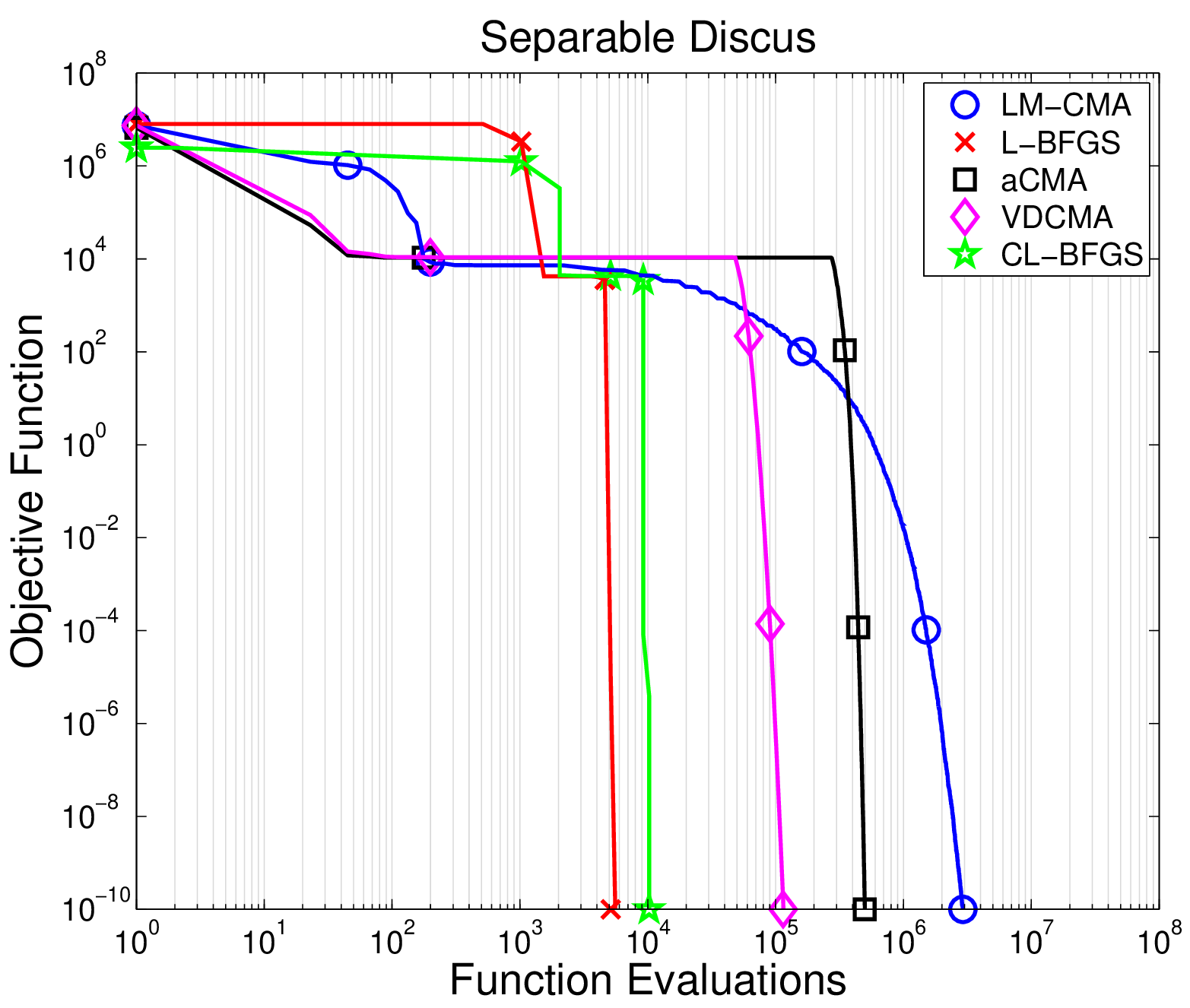}
  \includegraphics[width=0.5\textwidth]{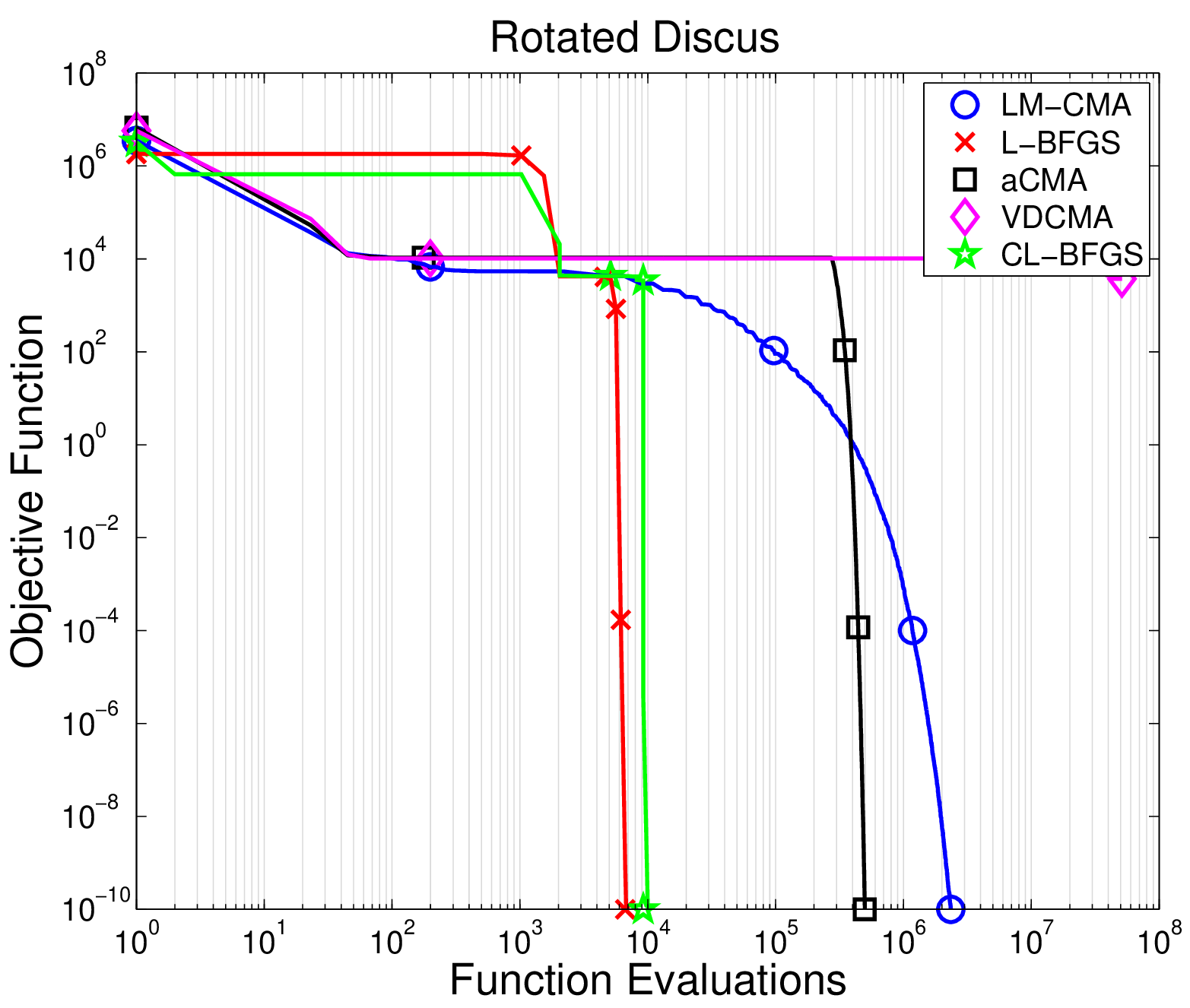}\\ 
	\includegraphics[width=0.5\textwidth]{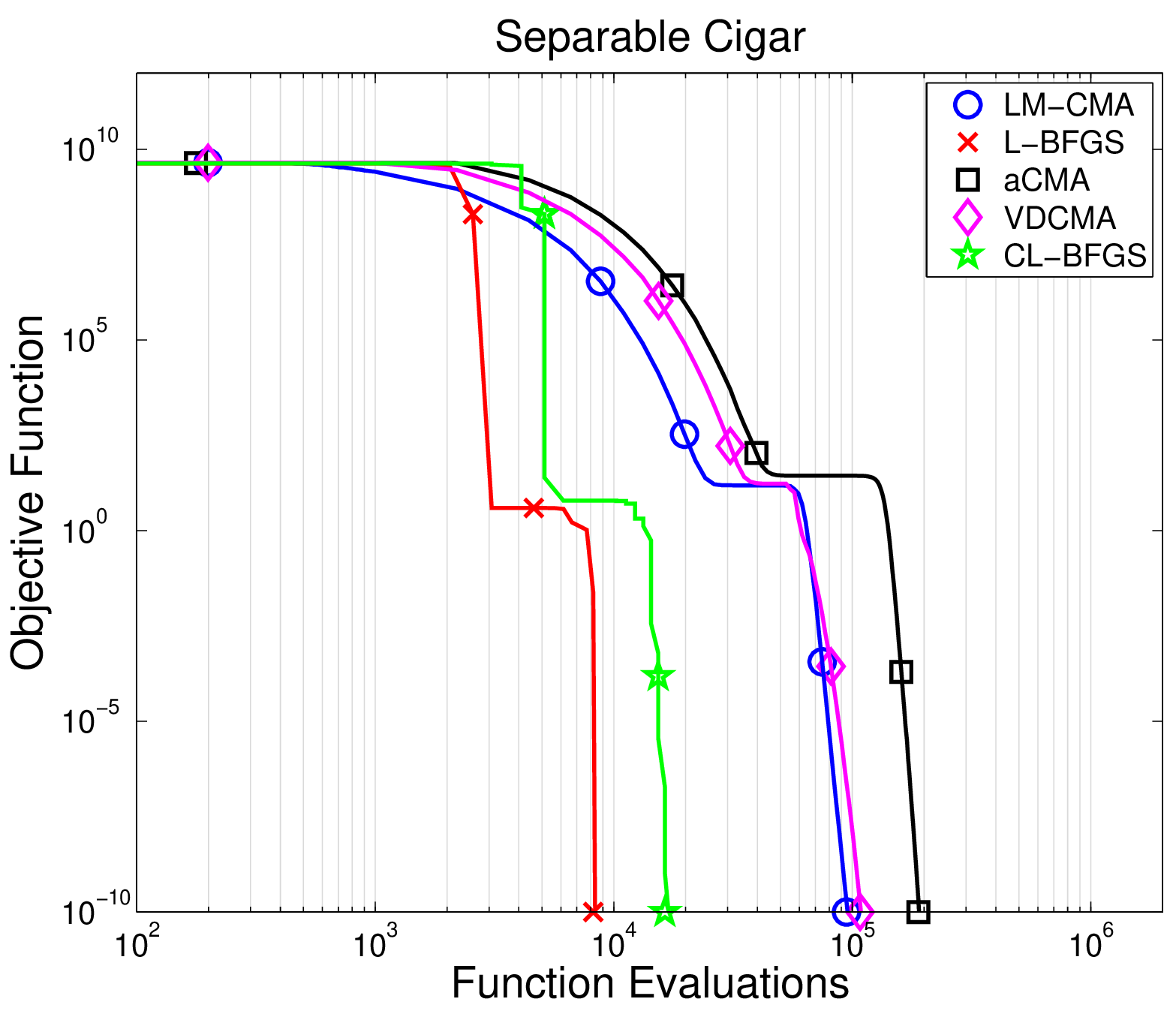}
  \includegraphics[width=0.5\textwidth]{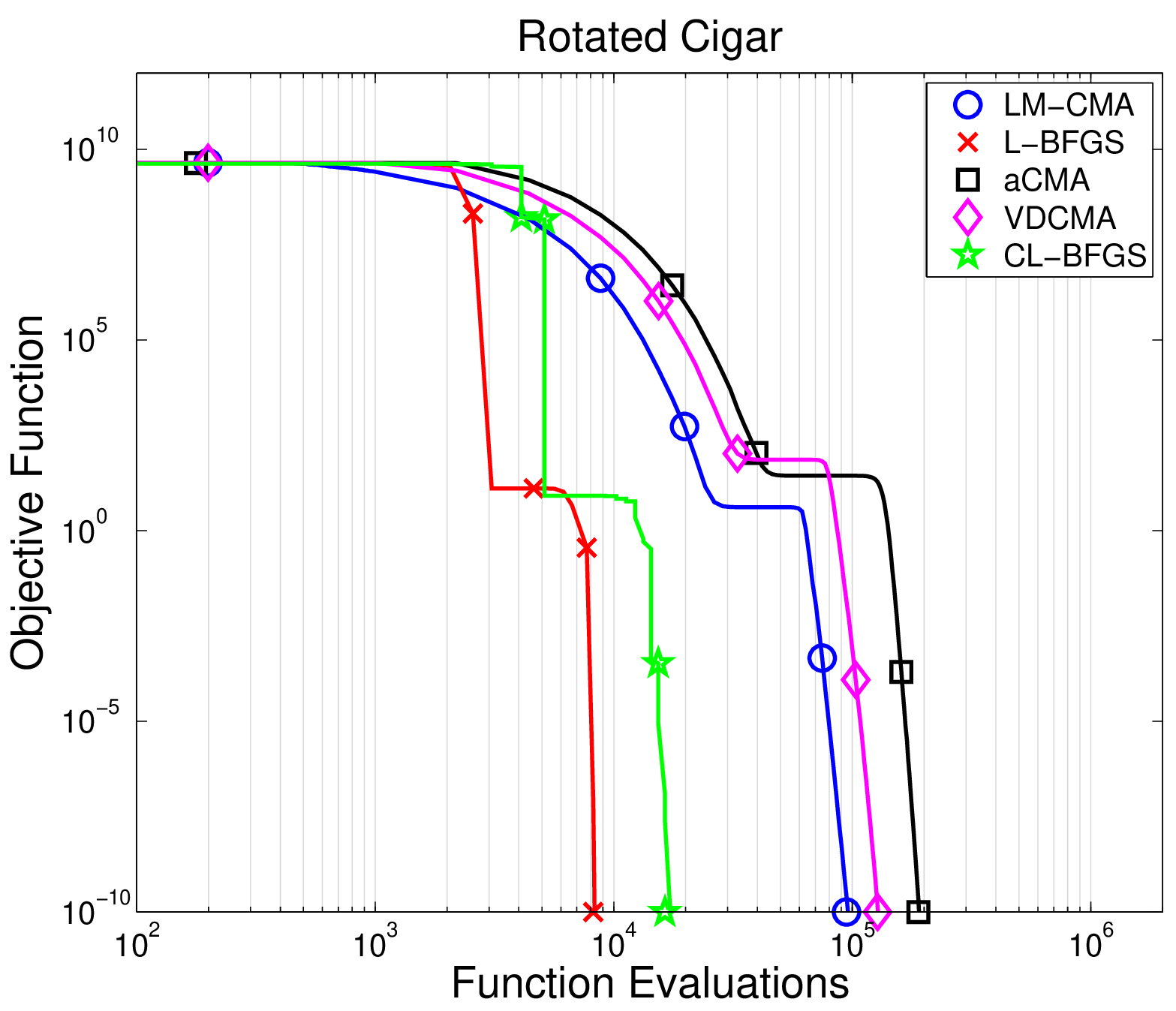}\\ 
	\includegraphics[width=0.5\textwidth]{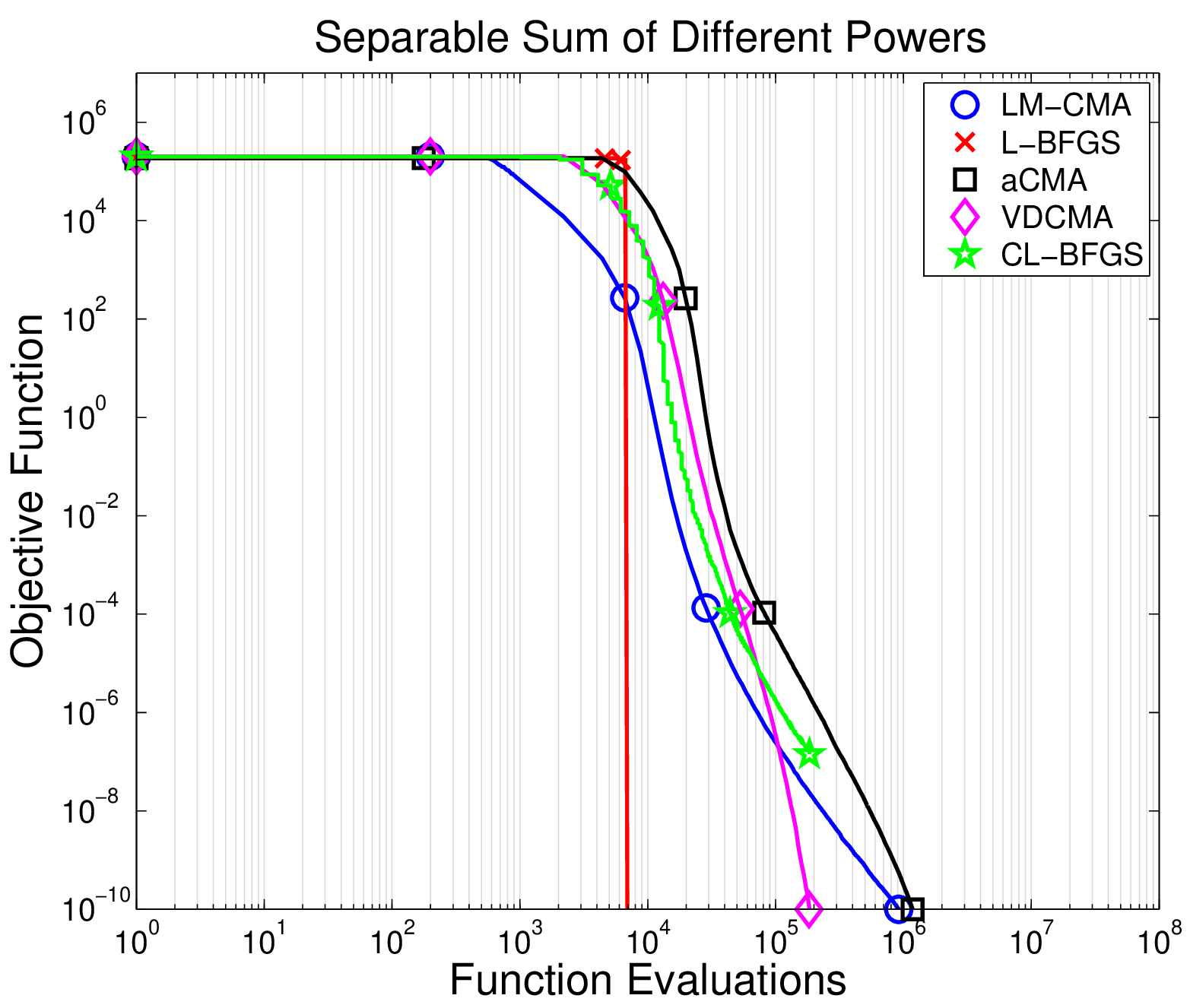}
  \includegraphics[width=0.5\textwidth]{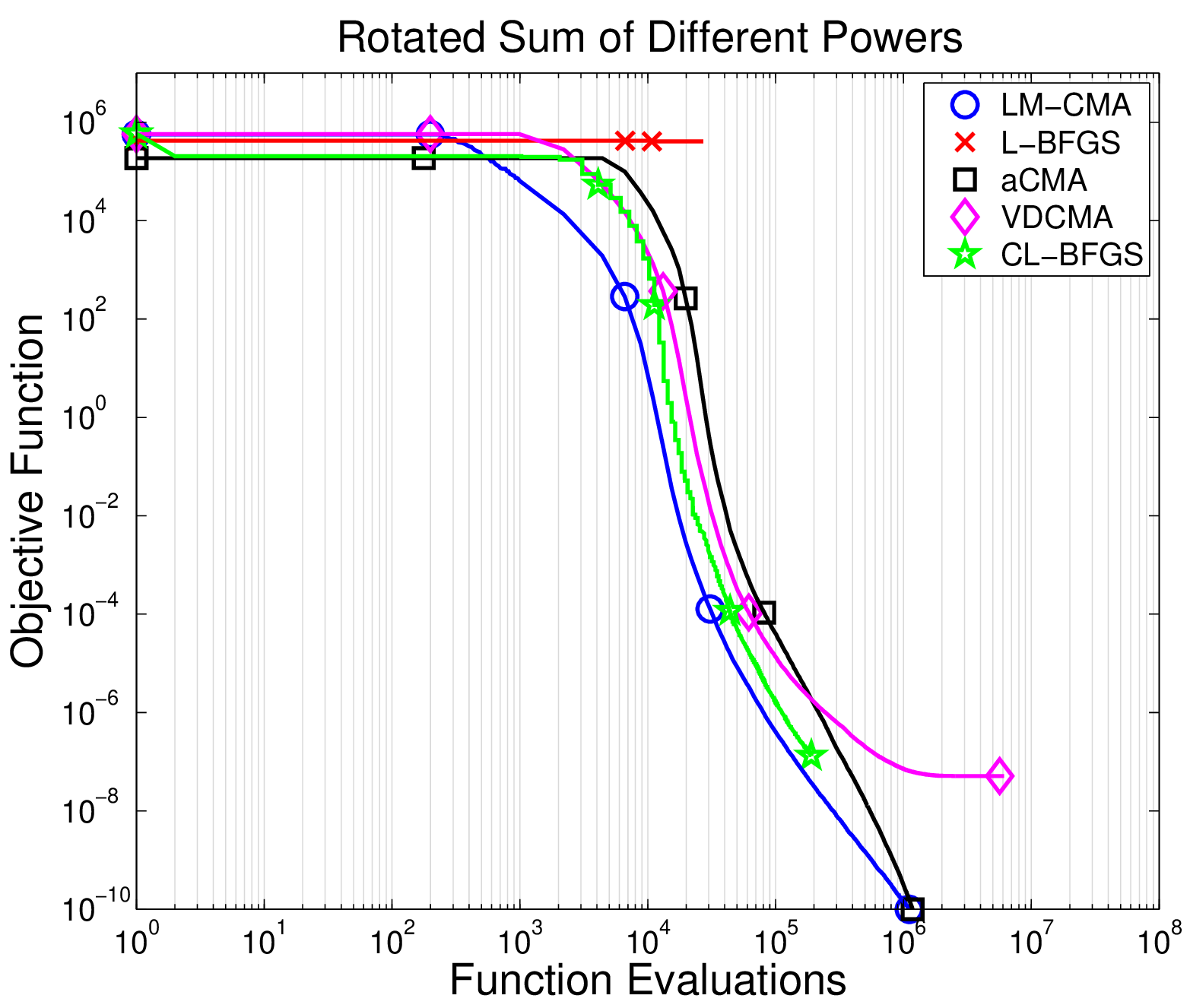}
\caption{\label{fig2} The trajectories show the median of 11 runs of LM-CMA, L-BFGS with exact gradients, CL-BFGS with central differencing, active CMA-ES and VD-CMA  on 512- Separable (Left Column) and Rotated (Right Column) functions.}
\end{figure}

Invariance properties under different search space transformations include translation invariance, scale invariance, rotational invariance and general linear invariance under any full rank matrix $\vc{R}$ when the algorithm performance on $f(\vc{x})$ and $f(\vc{R}\vc{x})$ is the same  given that the initial conditions of the algorithm are chosen appropriately \citep{2011HansenCMAandPSO}. 
 Thus, the lack of the latter invariance is associated with a better algorithm performance for some $\vc{R}$ and worse for the others. 
In practice, it often appears to be relatively simple to design an algorithm specifically for a set of problems with a particular $\vc{R}$, e.g., identity matrix, and then demonstrate its good performance. If this set contains separable problems, the problems where the optimum can be found with a coordinate-wise search, then even on highly multi-modal functions great results can be easily achieved \citep{2013LoshchilovHCMA}. Many  derivative-free search algorithms in one or another way exploit problem separability and fail to demonstrate a comparable performance on, e.g., rotated versions of the same problems. This would not be an issue if most of real-world problems are separable, this is, however, unlikely to be the case and some partial-separability or full non-separability are more likely to be present. 

\begin{figure}[!ht]
	\includegraphics[width=0.5\textwidth]{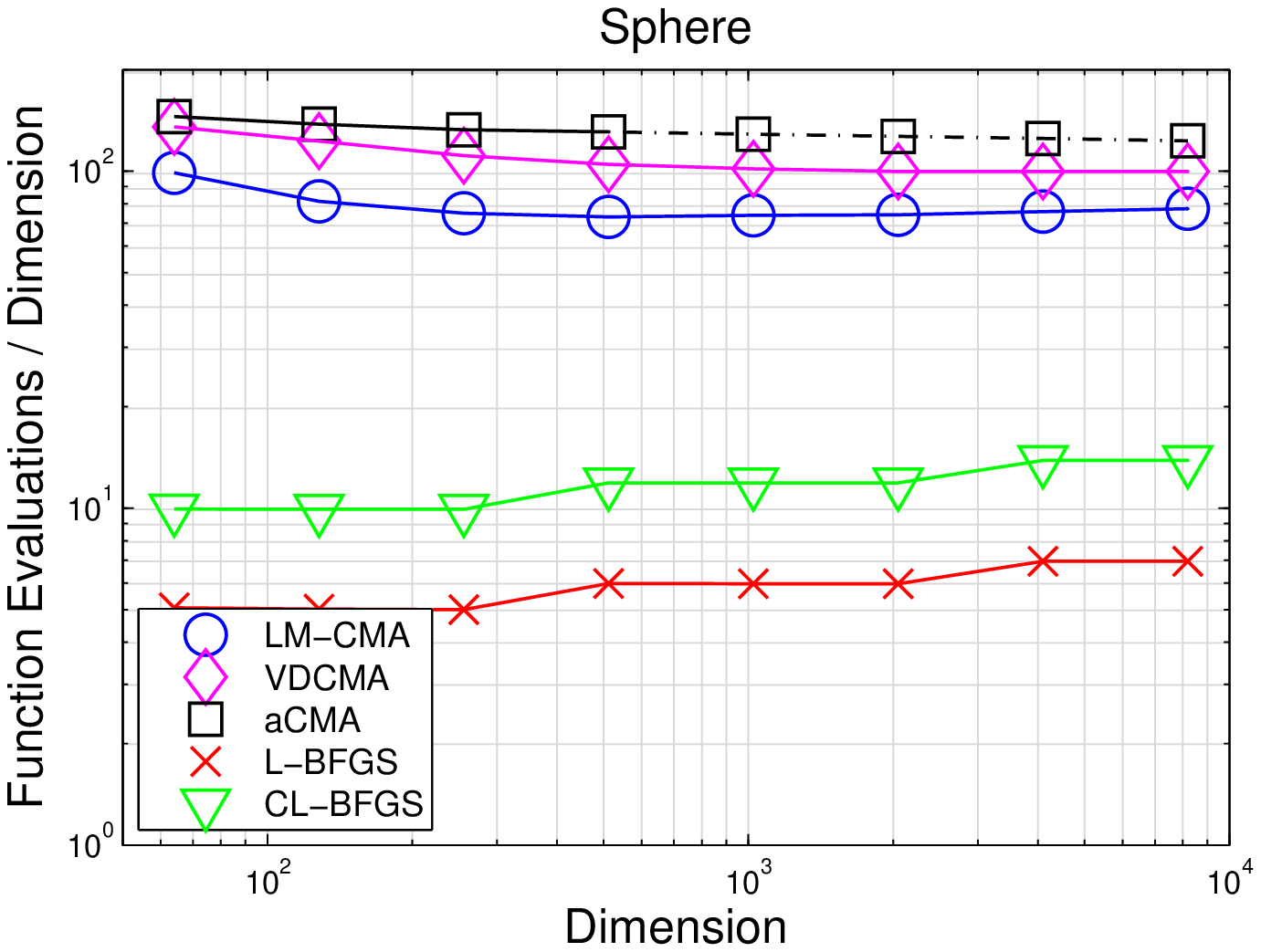}
  \includegraphics[width=0.5\textwidth]{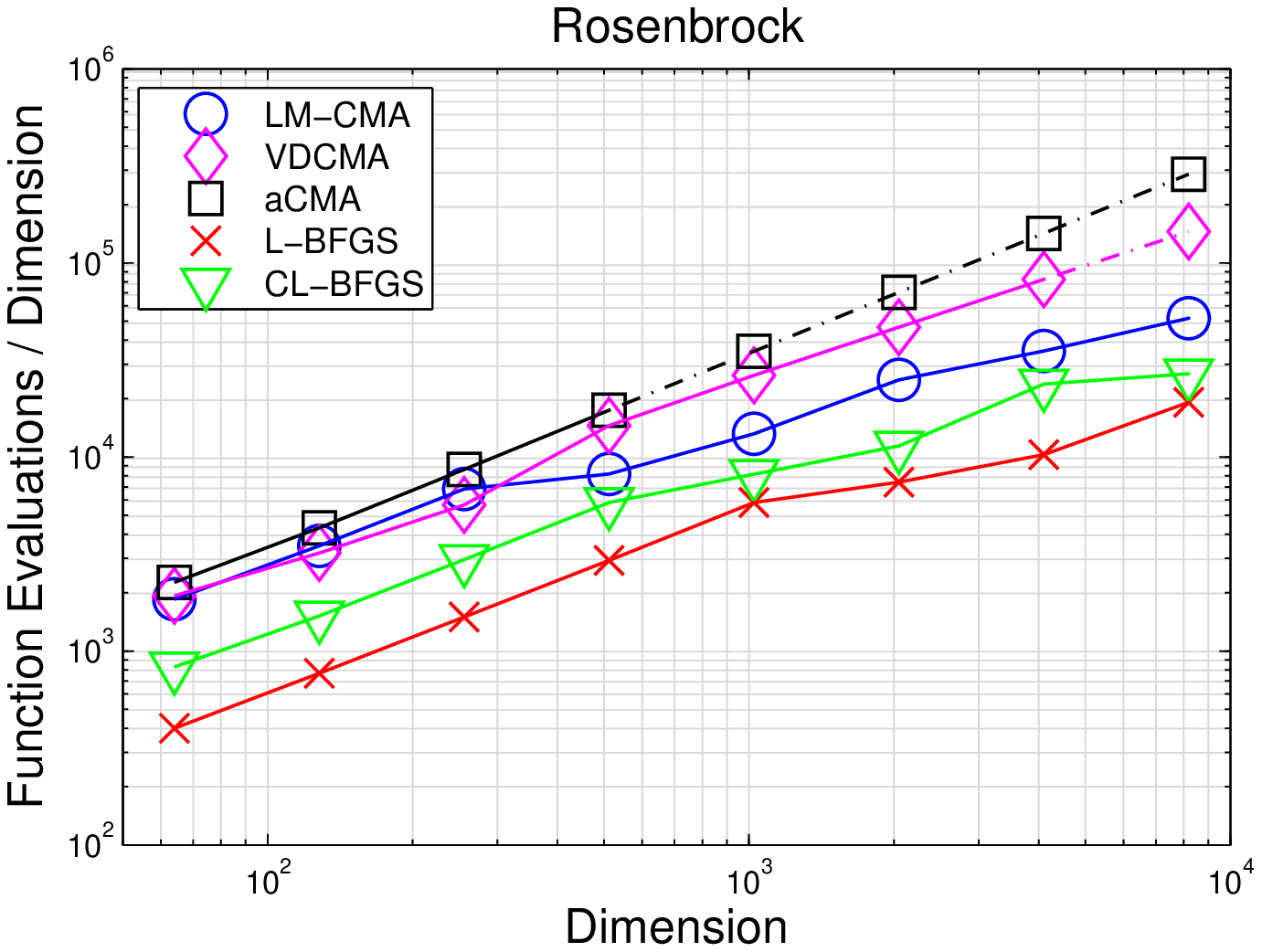}\\ 
	\includegraphics[width=0.5\textwidth]{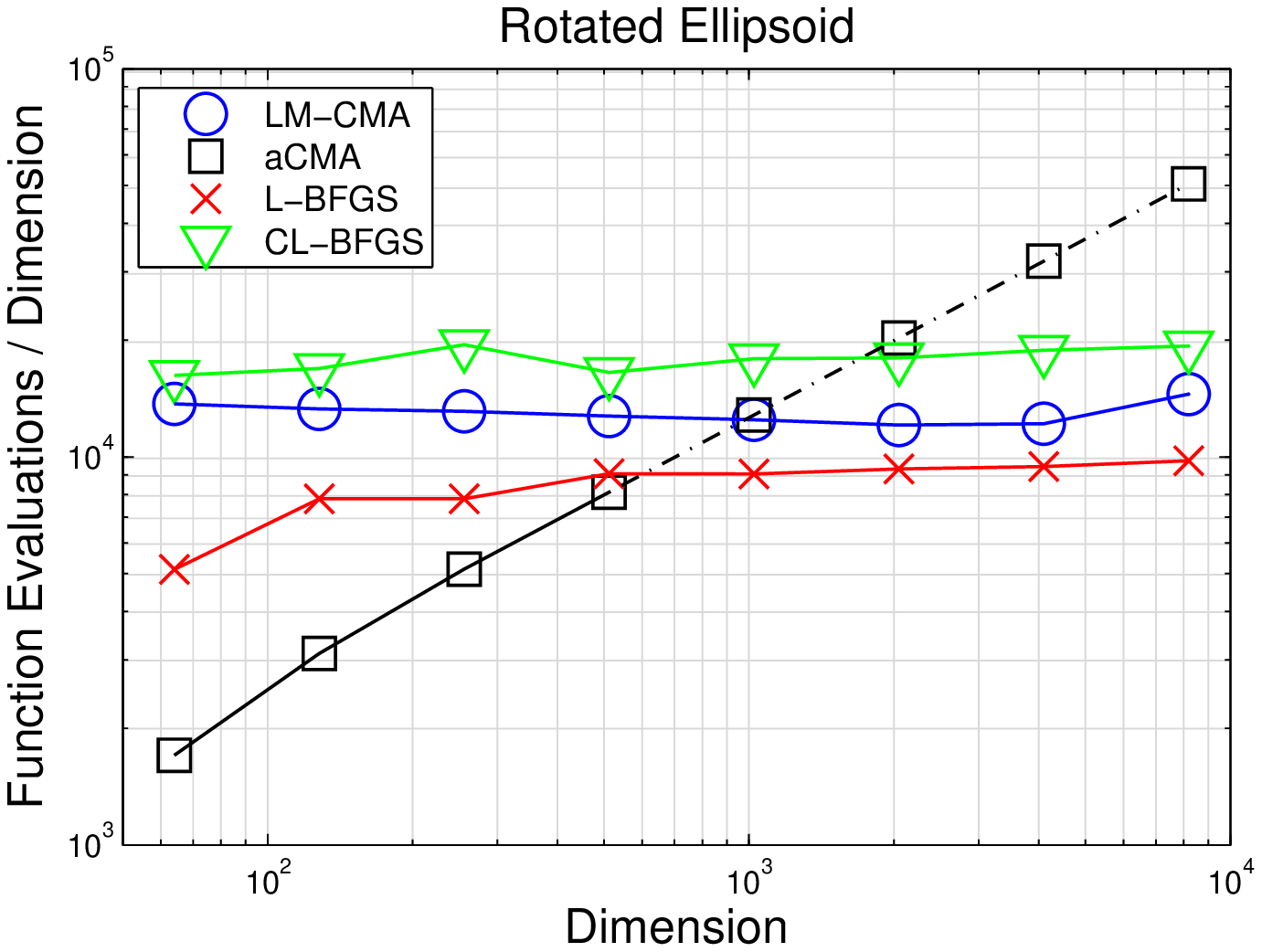}
  \includegraphics[width=0.5\textwidth]{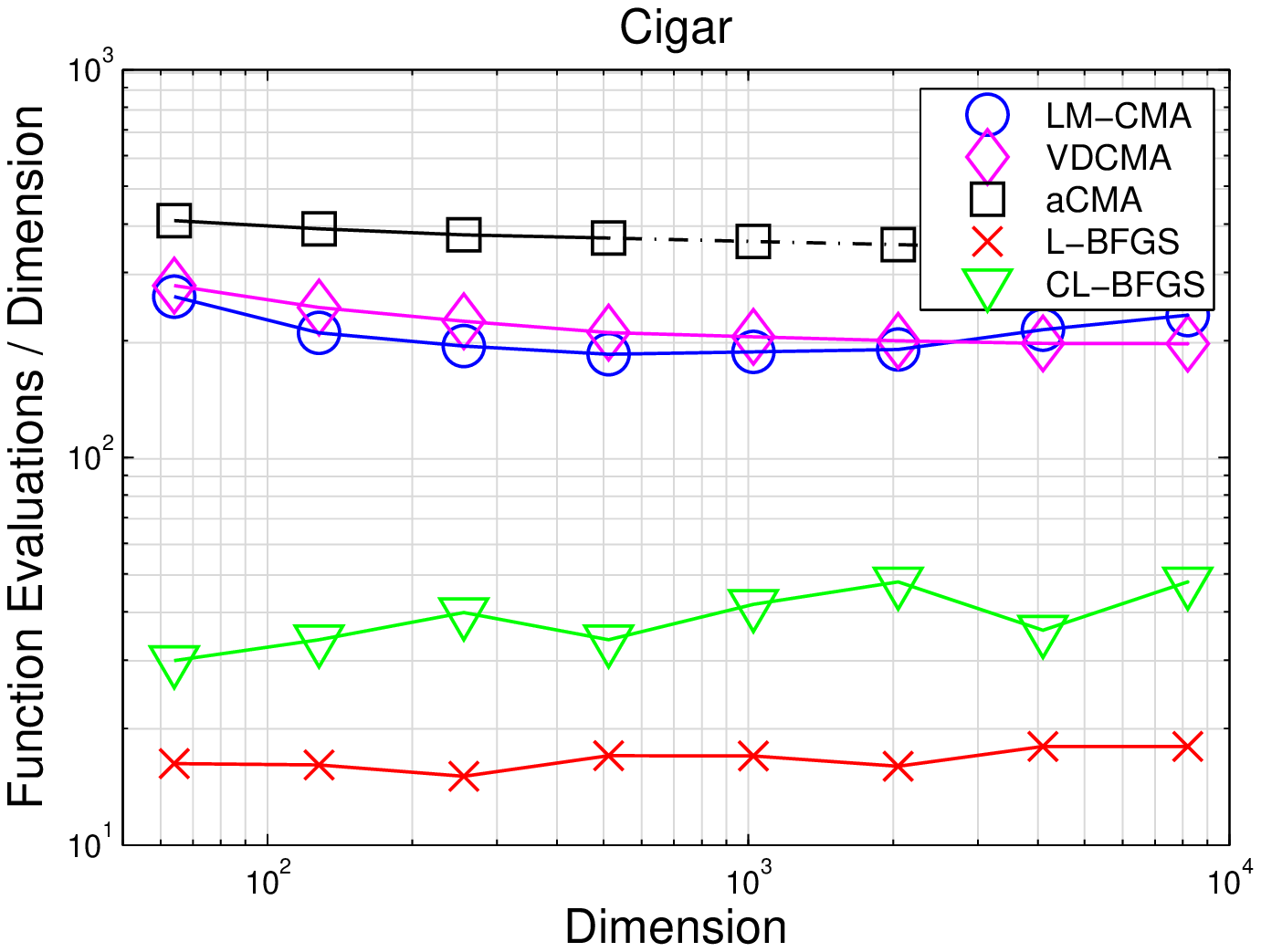}\\
	\includegraphics[width=0.5\textwidth]{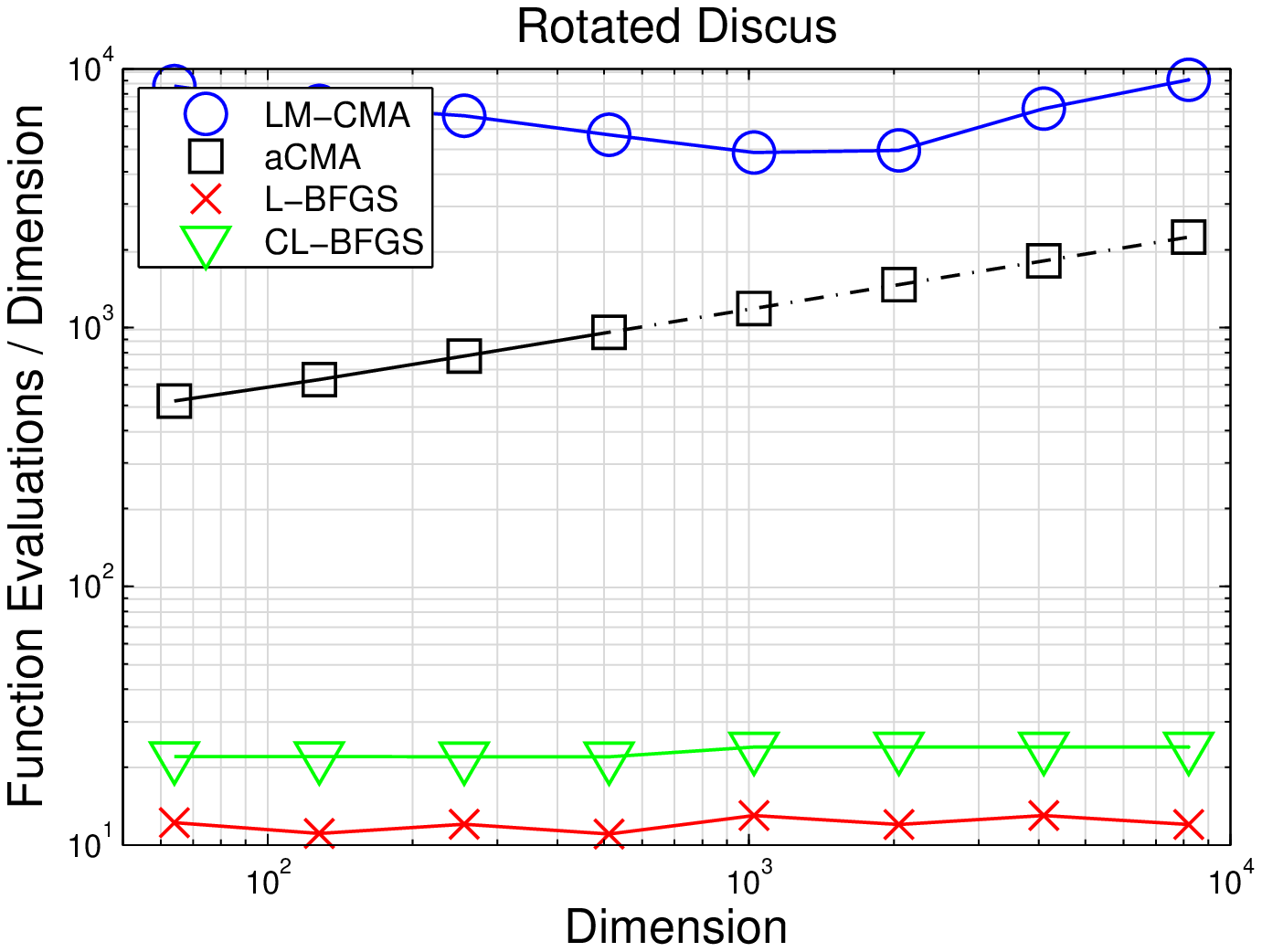}
  \includegraphics[width=0.5\textwidth]{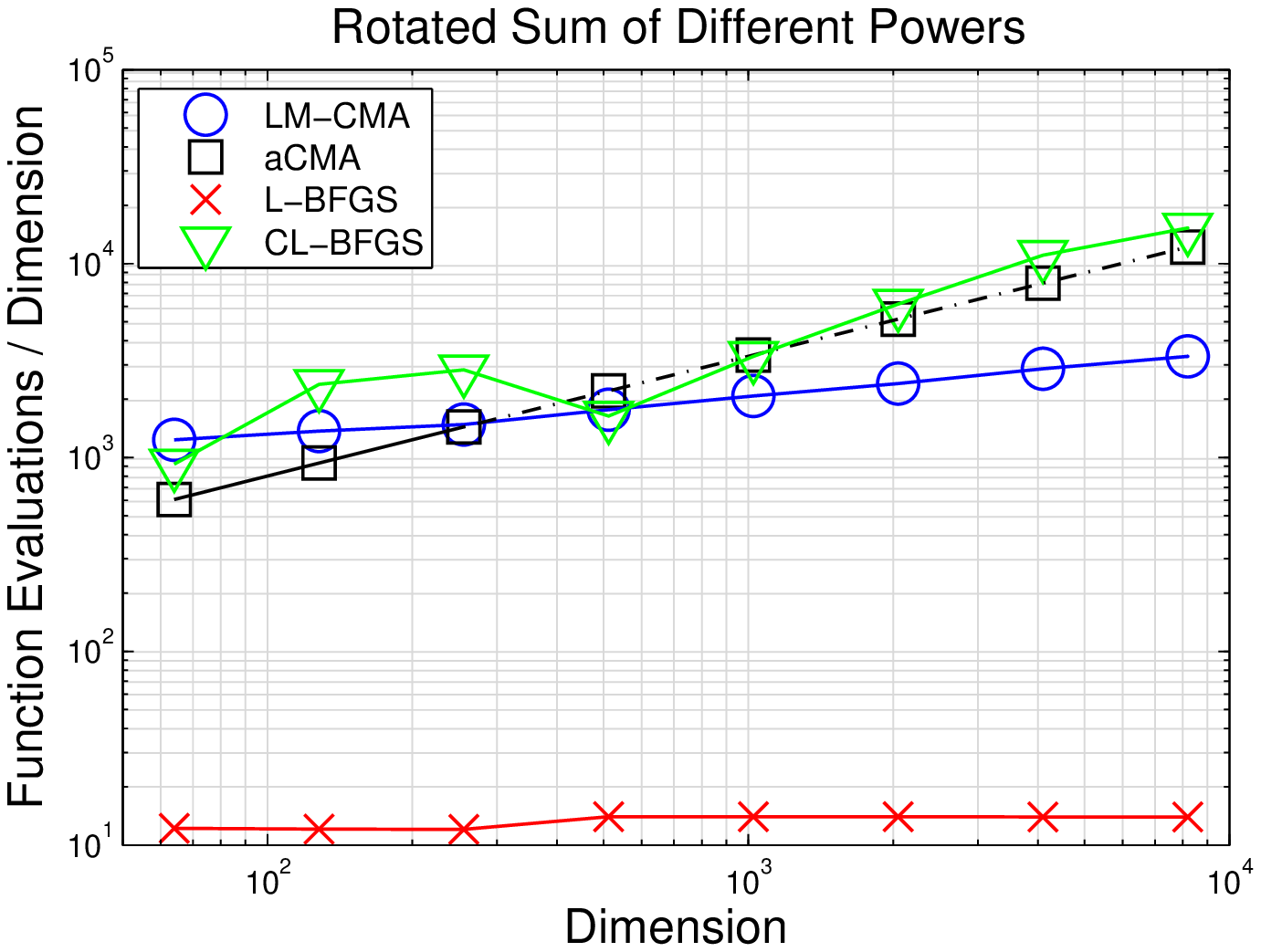}
\caption{\label{figvsdim} Median (out of 11 runs) number of function evaluations required to find $f(\vc{x})=10^{-10}$ for LM-CMA, L-BFGS with exact gradients, CL-BFGS with central differencing, active CMA-ES and VD-CMA. Dotted lines depict extrapolated results.}
\end{figure}

The original CMA-ES is invariant w.r.t. any invertible linear transformation of the search space, $\vc{R}$, if 
the initial covariance matrix $\C^{t=0} = \vc{R}^{-1} ( \vc{R}^{-1} )^{T}$, and the initial search point(s)
are transformed accordingly \cite{hansen2006cma}.  
 However, $\vc{R}$ matrix is often unknown 
(otherwise, one could directly transform the objective function) and cannot be stored in memory in large scale settings 
with $n \gg 10,000$. Thus, the covariance matrix adapted by LM-CMA has at most rank $m$ and so the intrinsic coordinate 
system cannot capture some full rank matrix $\vc{R}$ entirely.  Therefore,  
the performance of the algorithm on $f(\vc{R}\vc{x})$ compared to $f(\vc{x})$ depends on $\vc{R}$.  
However, in our experiments, differences in performance on axis-aligned and rotated ill-conditioned functions were marginal.

Here, we test LM-CMA, aCMA, L-BFGS, CL-BFGS both on separable problems and their rotated versions (see Table \ref{table:TableFunc}). 
It is simply intractable to run algorithms on large scale rotated problems with $n>1000$ due to the quadratic cost of involved matrix-vectors multiplications (see Figure \ref{fig:timing}). 
Fortunately, there is no need to do it for algorithms that are invariant to rotations of the search space since their performance is the same as on the separable problems whose evaluation is cheap (linear in time). 
Figures \ref{fig1}-\ref{fig2} show that the performance of aCMA on 512-dimensional (the dimensionality still feasible to perform full runs of aCMA) separable (left column) and rotated (right column) problems is very similar and the difference (if any) is likely due to a non-invariant initialization. The invariance in  performance is not guaranteed but rather observed for LM-CMA, L-BFGS and CL-BFGS. 
However, the performance of the VD-CMA degrades significantly on \rotelli, \rotdiscus\ and \rotdiffpow\ functions due to the restricted form of the adapted covariance matrix of Equation (\ref{VDCMAeq}). Both \rosen, \cigar\ and their rotated versions can be solved efficiently since they have a Hessian matrix whose inverse can be well approximated by Equation (\ref{VDCMAeq}) \citep{akimoto2014comparison}. 

An important observation from Figures \ref{fig1}-\ref{fig2} is that even the exact gradient information is not sufficient for L-BFGS to avoid numerical problems which lead to an imprecise estimation of the inverse Hessian matrix and premature convergence on \elli\ and \rotelli. The L-BFGS with the central difference method (CL-BFGS) experiences an early triggering of stopping criteria on \rosen\ and \diffpow. While numerical problems due to imprecise derivative estimations are quite natural for L-BFGS especially on ill-conditioned problems, we assume that with a better implementation of the algorithm (e.g., with high-precision arithmetic) one could obtain a more stable convergence. 
Therefore, we extrapolate the convergence curves of L-BFGS and CL-BFGS towards the target $f=10^{-10}$ after removing the part of the curve which clearly belongs to the stagnation, e.g., $f<10^{-7}$ on \elli.

\subsection{Scaling with Problem Dimension}
\label{sectscaling}

The performance versus the increasing number of problem variables is given in Figure \ref{figvsdim}. 
We exclude the results of VD-CMA on some problems because, as can be seen from Figures \ref{fig1}-\ref{fig2}, the algorithm does not find the optimum with a reasonable number of function evaluations or/and convergences prematurely. For algorithms demonstrating the same performance on separable and non-separable problems (see Figures \ref{fig1}-\ref{fig2}), we plot some results obtained on separable problems as obtained on rotated problems in Figure \ref{figvsdim} to avoid possible misunderstanding from designers of separability-oriented algorithms.

The results suggest that L-BFGS shows the best performance, this is not surprising given the form of the selected objective functions (see Table \ref{table:TableFunc}). We should also keep in mind that the exact gradients were provided and this still led to premature convergence on some functions (see Figures \ref{fig1}-\ref{fig2}). In the black-box scenario, one would probably use L-BFGS with the forward or central (CL-BFGS) difference methods. The latter is often found to lead to a loss by a factor of 2 (as expected due to $2n+1$ versus $n+1$ cost per gradient), except for the \rotdiffpow, where the loss is increasing with problem dimension.

Quite surprisingly, the LM-CMA outperforms VD-CMA and aCMA on \sphere. This performance is close to the one obtained for (1+1) Evolution Strategy with optimal step-size. Bad performance on \sphere\ anyway would not directly mean that an algorithm is useless, but could illustrate its performance in vicinity of local optima when variable-metric algorithms (e.g., CMA-like algorithms) may perform an isotropic search w.r.t. an adapted internal coordinate system. The obtained results are mainly due to the Population Success Rule which deserves an independent study similar to the one by \cite{hansen2014assess}. 
Nevertheless, we would like to mention a few key points of the PSR. 
By design, depending on the target success ratio $z^*$, one can get either biased (for $z^*\neq 0$) or unbiased (for $z^*= 0$) random walk on random functions. It would be a bias to say that either biased or unbiased change of $\sigma$ "is better" on random functions, since the latter depends on the context. Due to the fact that the (weighted) mean of each new population is computed from the best $\mu$ out of $\lambda$ individuals, the $\lambda$ individuals of the new generation are typically as good as the (weighted) best $\mu$ individuals of the previous one, and, thus, if $z^*=0$ one may expect $z_{PSR}>0$ from Equation (\ref{ZPSR}). Typically, it is reasonable to choose $z^*\in(0,0.5)$ lower-bounded by 0 due to random functions and upper-bounded by 0.5 due to linear functions. In this study, we choose 0.3 which lies roughly in the middle of the interval. It is important to mention a striking similarity with the 1/5th success rule \citep{schumer1968adaptive,1973RechenbergEvolutionsstrategie}. We consider the PSR to be its population-based version.

\begin{figure}[!ht]
	\includegraphics[width=0.5\textwidth]{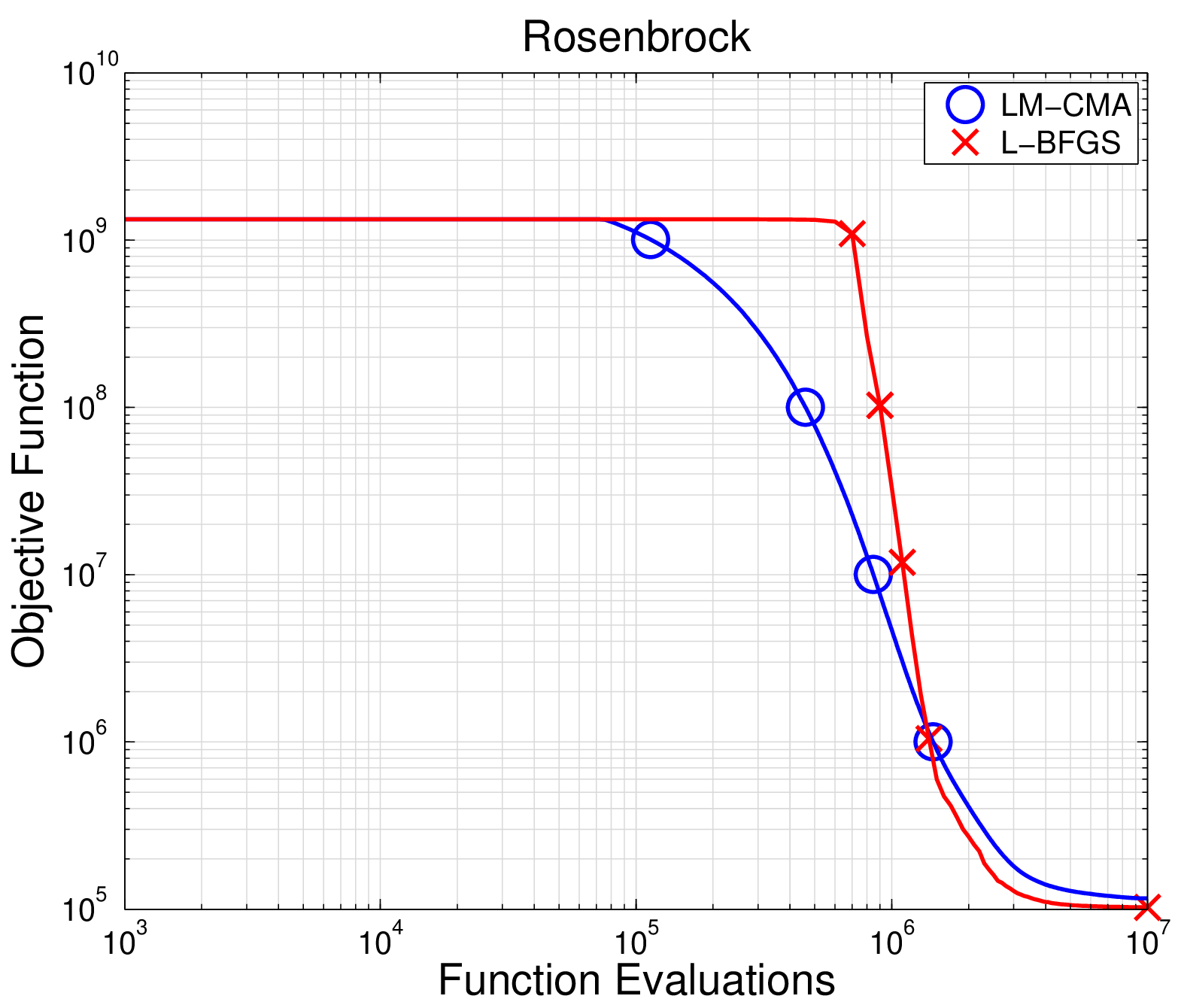}
  \includegraphics[width=0.5\textwidth]{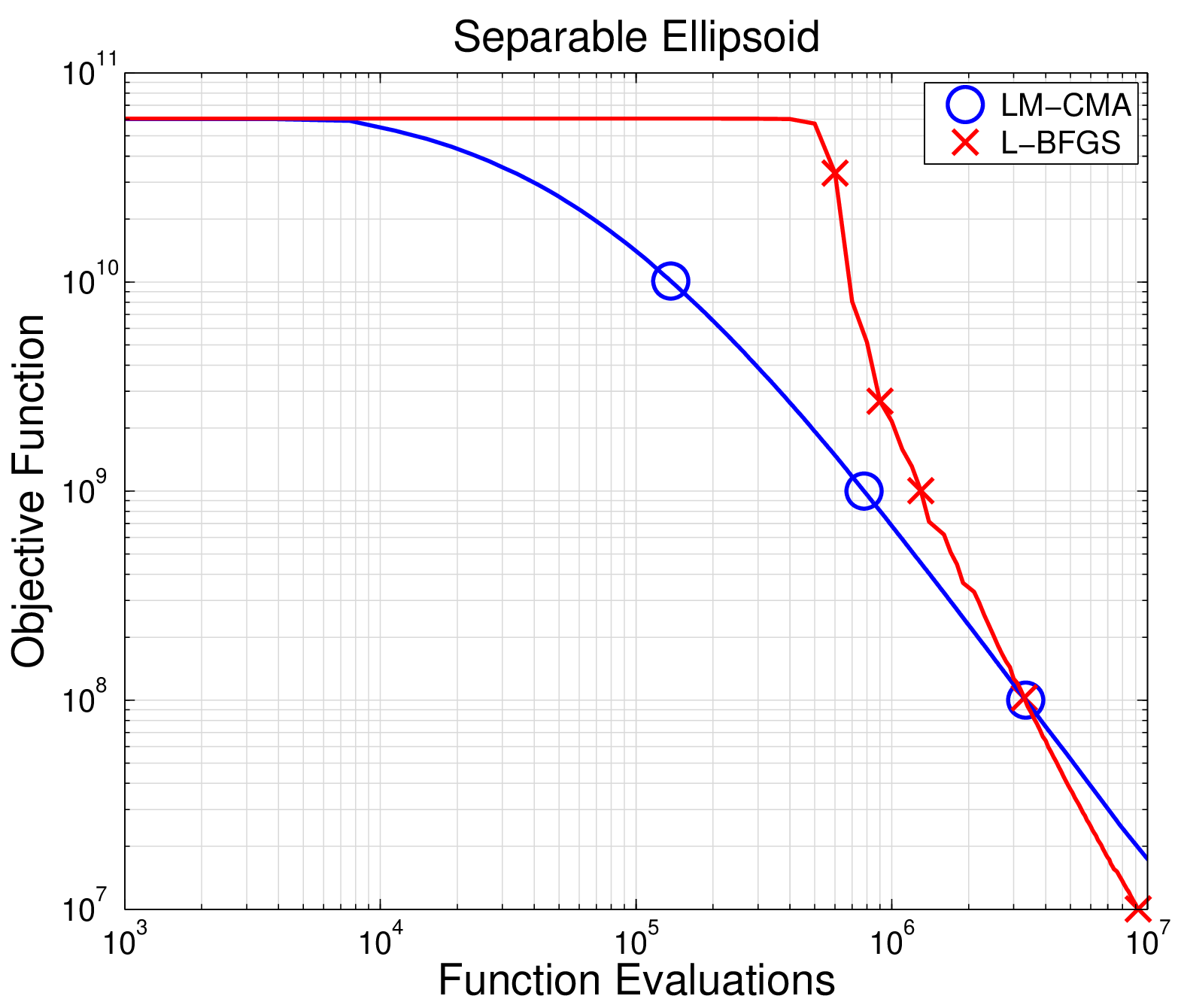}
\caption{\label{fig100k} The trajectories show the median of 11 runs of LM-CMA, L-BFGS (with exact gradients provided at the cost of $n+1$ evaluations per gradient) on 100,000-dimensional Rosenbrock and Ellipsoid functions.}
\end{figure}

 The performance of LM-CMA on \elli\ is probably the most surprising result of this work. 
In general, the scaling of CMA-ES is expected to be from super-linear to quadratic with $n$ on \elli\ since the number of parameters of the full covariance matrix to learn is $(n^2+n)/2$ \citep{2001HansenCMA}. While aCMA  demonstrates this scaling, LM-CMA scales linearly albeit with a significant constant factor. The performance of both algorithms coincides at $n\approx 1000$, then, LM-CMA outperforms aCMA (given that our extrapolation is reasonable) with a factor increasing with $n$. It should be noted that aCMA is slower in terms of CPU time per function evaluation by a factor of $n/10$ (see Figure \ref{fig:timing}). Another interesting observation is that the L-BFGS is only slightly faster than LM-CMA, while CL-BFGS is actually outperformed by the latter. An insight to these observations can be found in Figure \ref{fig1} where both LM-CMA and L-BFGS outperform aCMA by a factor of 10 in the initial part of the search, while aCMA compensates this loss by having the covariance matrix well adapted that allows to accelerate convergence close to the optimum. This might be explained as follows: a smaller number of internal parameters defining the intrinsic coordinate system can be learned faster and with greater learning rates, this allows a faster convergence but may slow down the search in vicinity of the optimum if the condition number cannot be captured by the reduced intrinsic coordinate system. 

The LM-CMA is better or is as good as VD-CMA on \rosen\ and \cigar\ where it is expected to be outperformed by the latter due a presumably few principal components needed to be learnt to solve these problems. The scaling on \rosen\ suggests that the problem is more difficult (e.g., more difficult than \elli) than one could expect, mainly due to an adaptation of the intrinsic coordinate system required while following the banana shape valley of this function.

The results on 100,000-dimensional problems (see Figure \ref{fig100k}) show that LM-CMA outperforms L-BFGS  on the first $10n - 20n$ function evaluations which corresponds to the first 10-20 iterations of L-BFGS. This observation suggests that LM-CMA can be viewed as an alternative to L-BFGS when $n$ is large and the available number of function evaluations is limited. While it can provide a competitive performance in the beginning, it is also able to learn dependencies between variables to approach the optimum.

\begin{figure}[t]
	\includegraphics[width=0.5\textwidth]{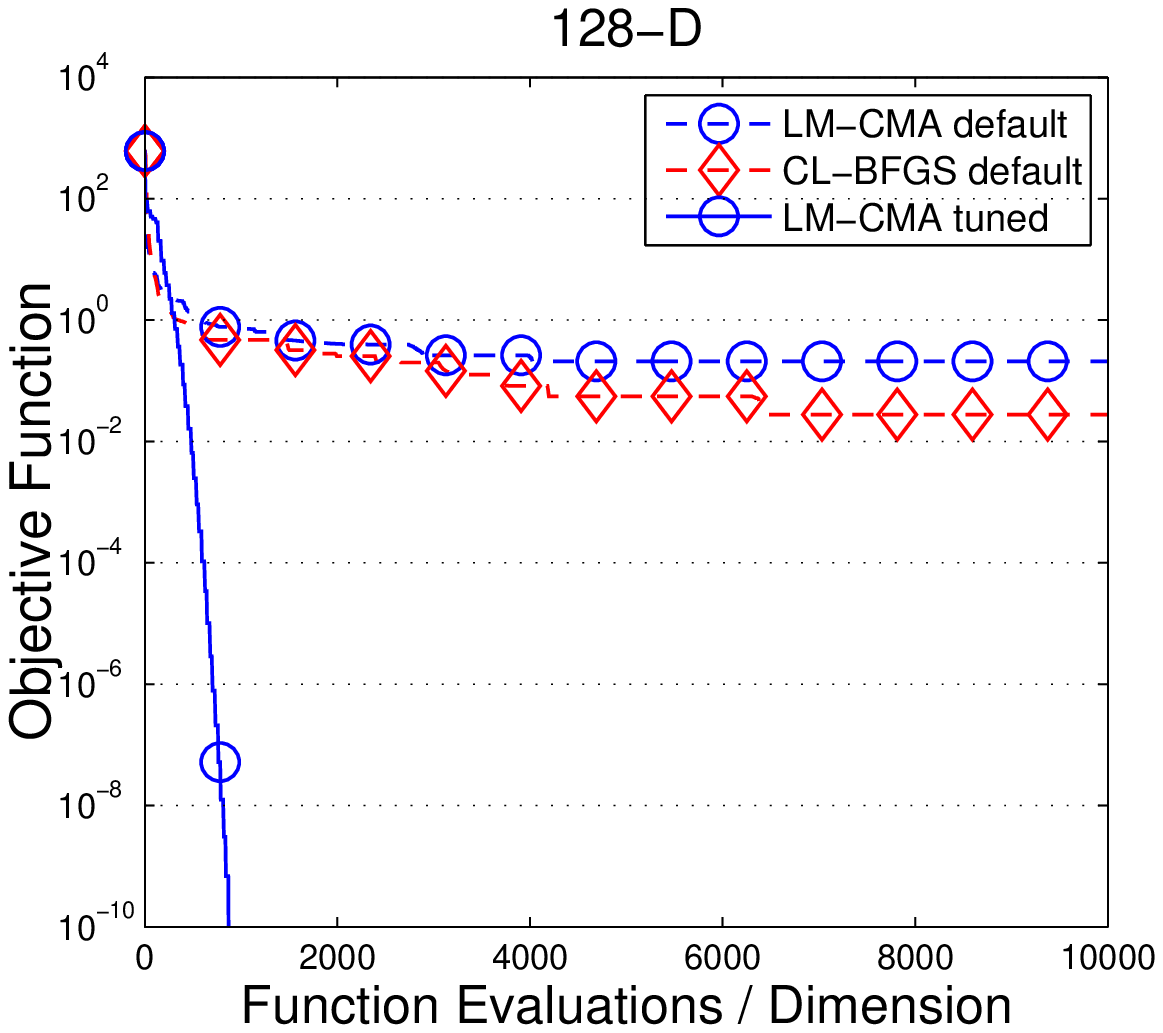}
	\includegraphics[width=0.5\textwidth]{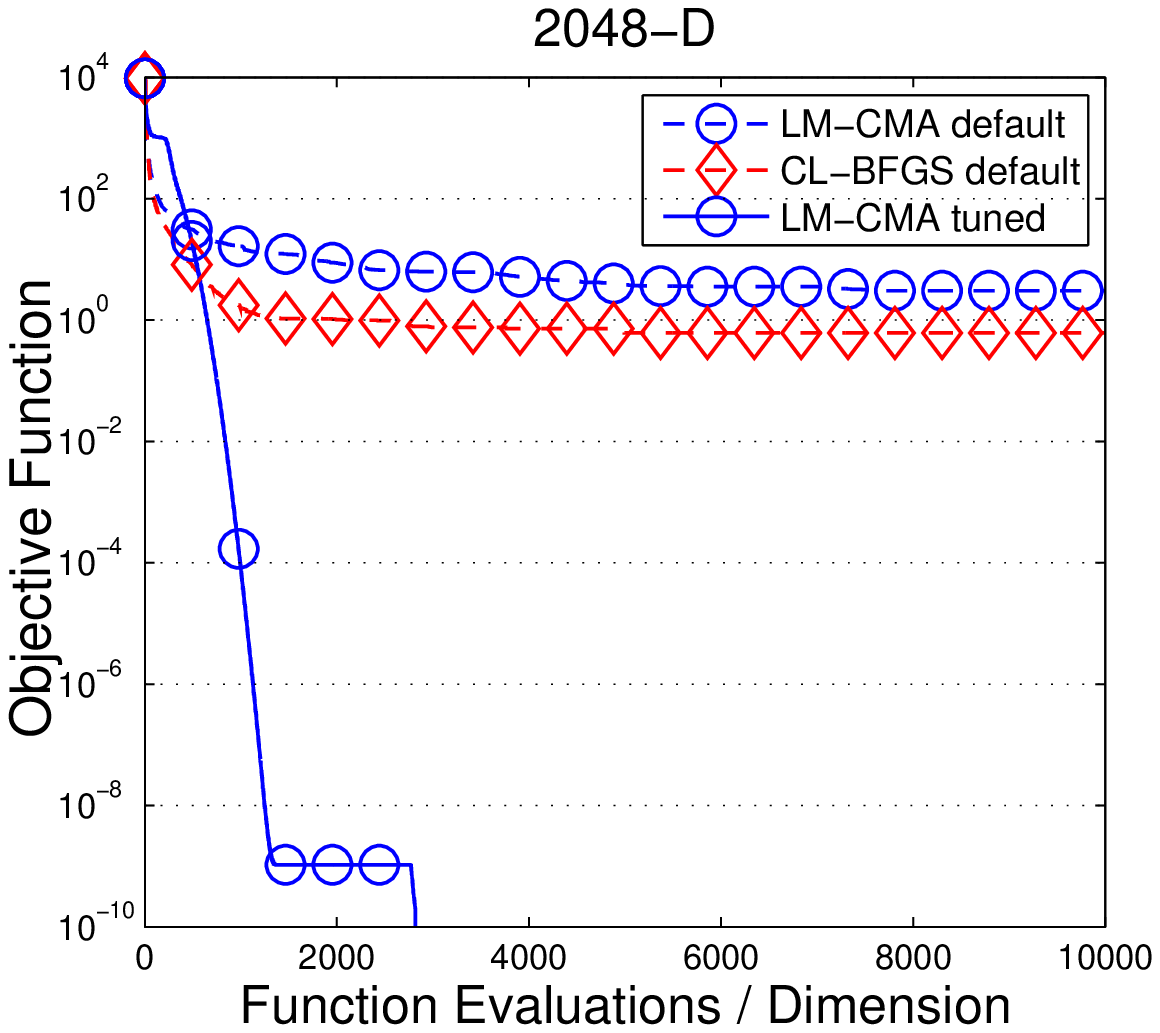}
\caption{\label{figNest} The trajectories show the median of 11 runs of LM-CMA in default settings, 
CL-BFGS in default settings and tuned LM-CMA (all three algorithms are with restarts) 
on the second nonsmooth variant of Nesterov-Chebyshev-Rosenbrock function in dimensions 128 and 2048.}
\end{figure}

\begin{figure}[t]
	\includegraphics[width=0.5\textwidth]{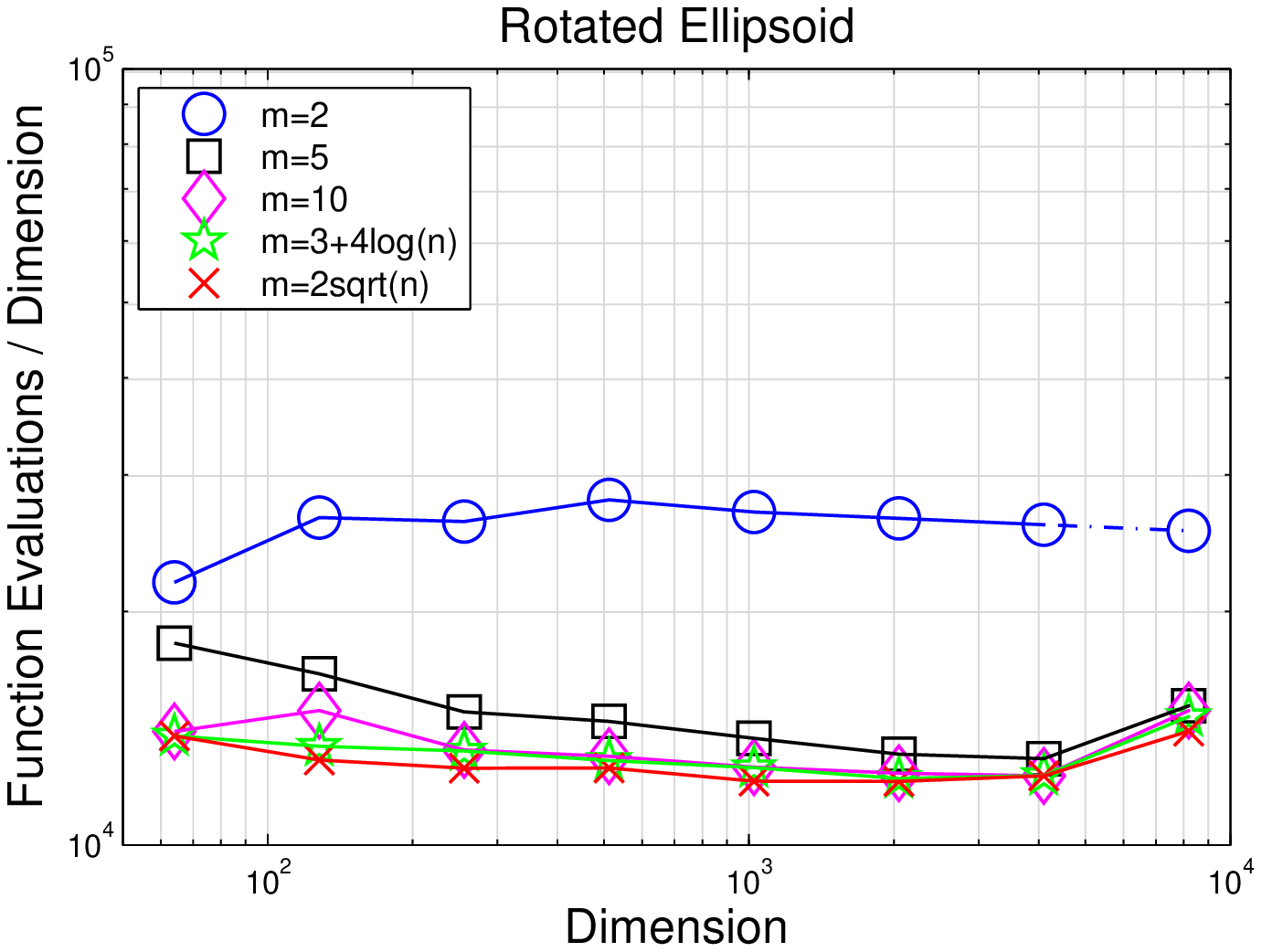}
  \includegraphics[width=0.5\textwidth]{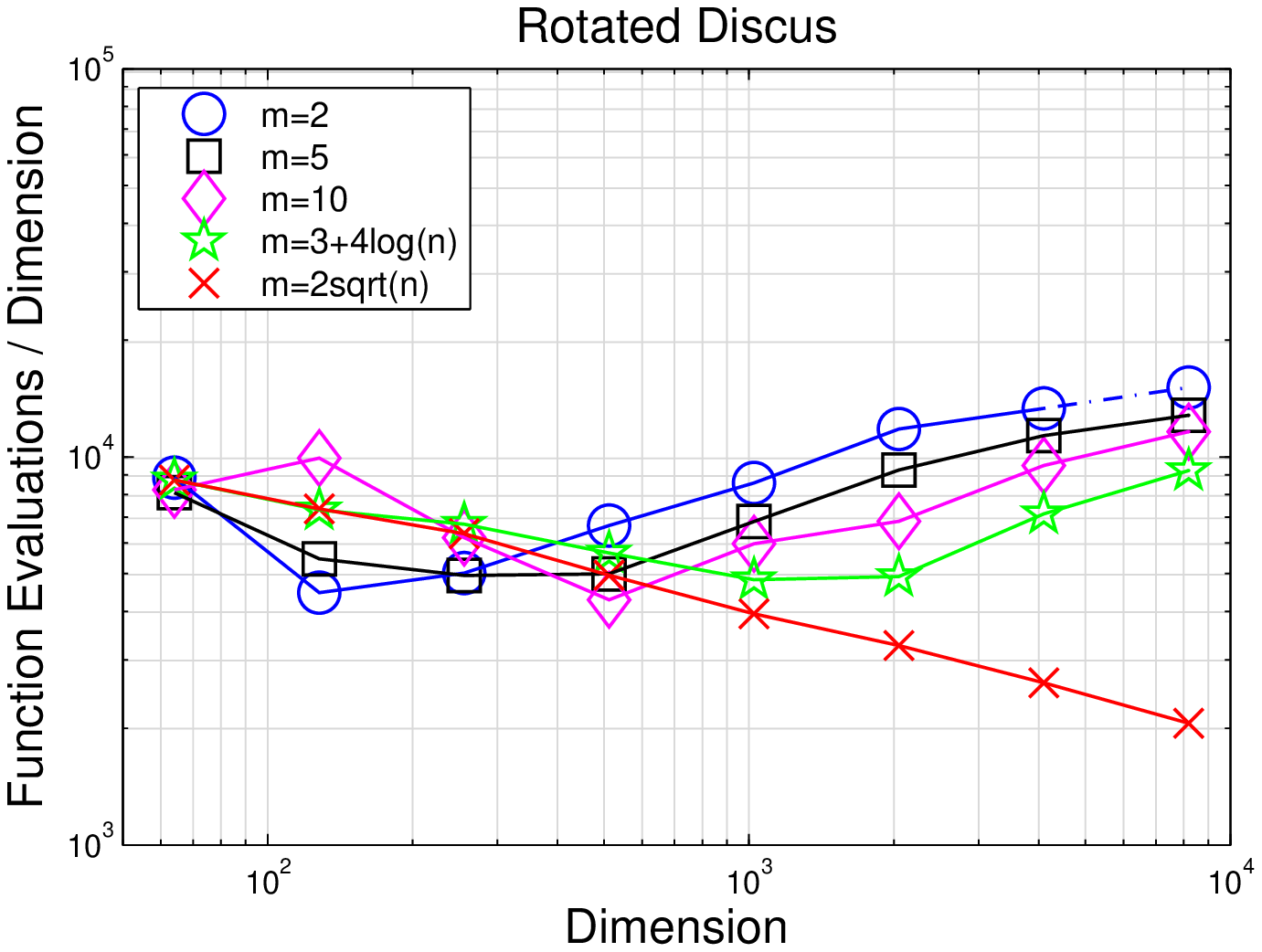}\\ 
	\includegraphics[width=0.5\textwidth]{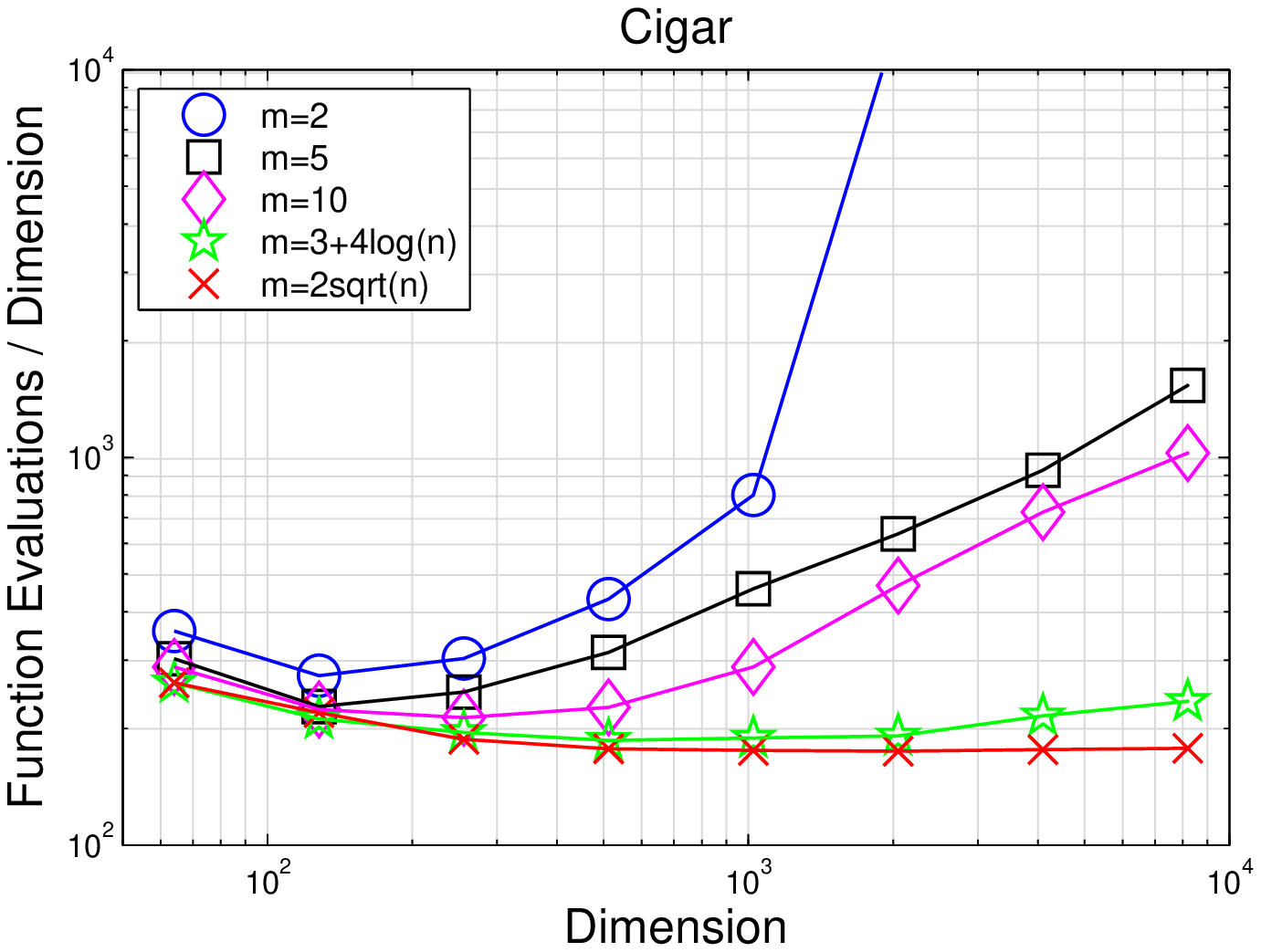}
  \includegraphics[width=0.5\textwidth]{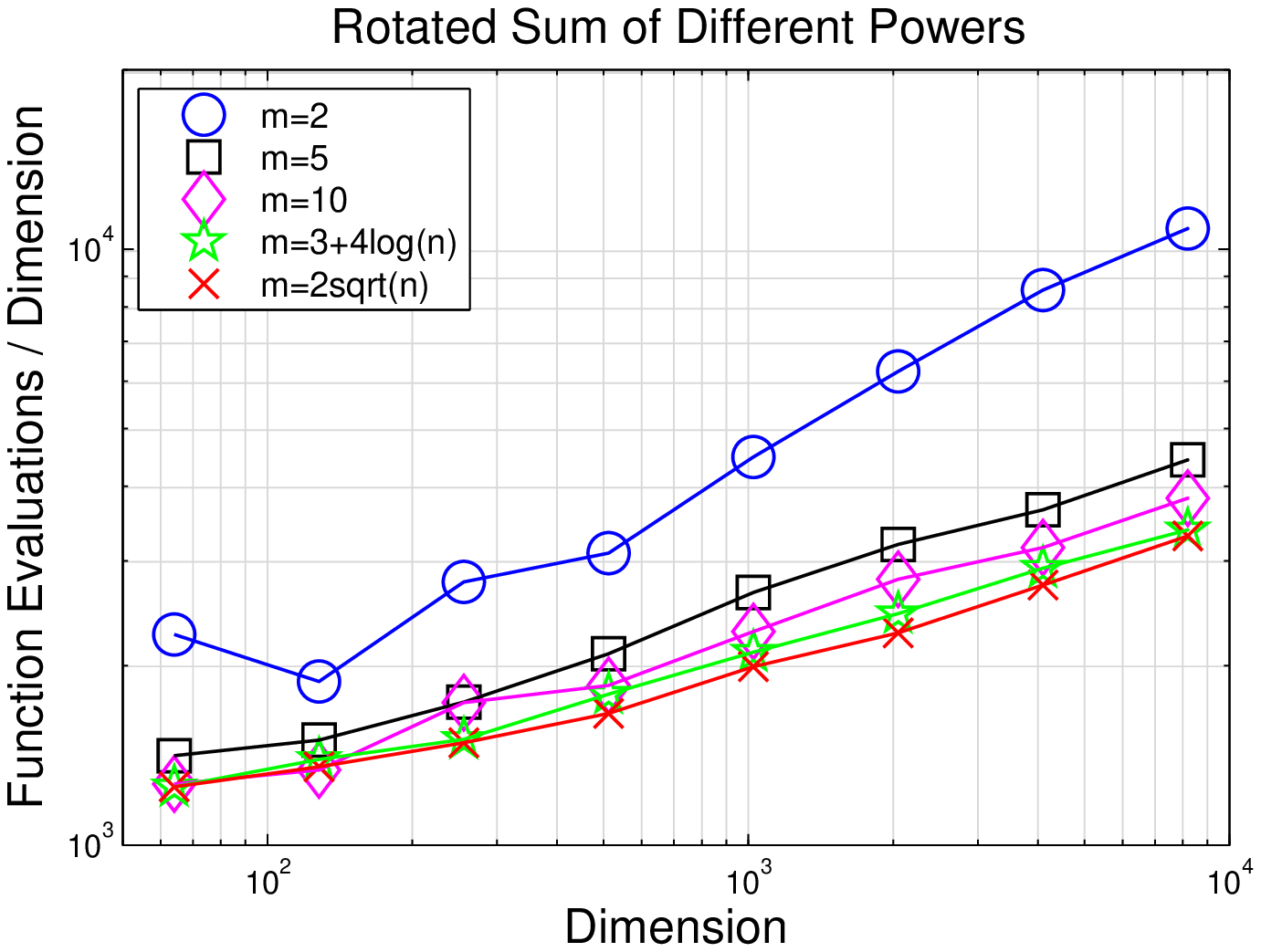}
\caption{\label{figMsens} Sensitivity of LM-CMA to different settings of $m$.}
\end{figure}

\subsection{Performance on a nonsmooth variant of Nesterov's function}
\label{secnest}

While designed for smooth optimization, 
BFGS is known to work well for nonsmooth optimization too. 
A recent study by \cite{Overton2015} demonstrated 
the difficulties encountered by BFGS on some nonsmooth 
functions. We selected one of the test functions from \cite{Overton2015} called 
the second nonsmooth variant of Nesterov-Chebyshev-Rosenbrock 
function defined as follows:

	\begin{equation} 
	\label{Fnest}
  \hat{N}(x) = \frac{1}{4} \left| x_1 - 1 \right| +
	\sum_{i=1}^{n-1} \left| x_{i+1} - 2\left| x_i \right| + 1\right|
  \end{equation}

This function is nonsmooth (though locally Lipschitz) as well as
nonconvex, it has $2^{n-1}$ Clarke stationary points \citep{Overton2015}. 
\cite{Overton2015} showed that for $n=5$ BFGS starting from 1000 
randomly generated points finds all 16 Clarke stationary points (for the definition of Clarke stationary points see \cite{abramson2006convergence}) and the probability to find the global minimizer is only by about a factor of 2 greater than to find any of the  Clarke 
stationary points. This probability dropped by a factor of 2 for $n=6$ while and since the number of Clarke stationary points doubled \citep{Overton2015}. Clearly, the problem becomes extremely difficult for BFGS when $n$ is large. 

We launched LM-CMA and CL-BFGS (L-BFGS performed worse) on $\hat{N}(x)$ for $n=128$ and $n=2048$. 
Figure \ref{figNest} shows that CL-BFGS performs better than LM-CMA, 
however, both algorithms in default settings and with restarts do not perform well. 
We tuned both LM-CMA and CL-BFGS but report the results only for LM-CMA since we failed to improve 
the performance of CL-BFGS by more than one order of magnitude of the objective function value. 
The tuned parameters for LM-CMA are: i) doubled population size $\lambda$, 
ii) increased learning rate by 15 to $c_1 = 15/(10\ln(n+1))$, iii) an extremely small learning rate for step-size adaptation  
$c_{\sigma} = 0.3/n^2$ instead of $c_{\sigma} = 0.3$. 
The last modification is probably the most important, practically, it defines the schedule how step-size decreases. 
A similar effect can be achieved by reducing $z^{*}$ or increasing $d_{\sigma}$. 
Faster learning of dependencies between variables and slower step-size decrease drastically improve the convergence 
and the problem can be solved both for $n=128$ and $n=2048$ (Figure \ref{figNest}). 
Interestingly, the number of function evaluations scales almost linearly with problem dimension.

We expected that tuning of CL-BFGS will lead to similar improvements. Surprisingly, 
our attempts to modify its parameters, often in order to slow down the convergence (e.g., type and number of line-search steps, Wolfe conditions parameters) failed. We still expect that certain modifications should improve CL-BFGS and thus we leave this question open.   
The settings tuned for $\hat{N}(x)$ function differ significantly from the default ones. 
It is of great interest to find an online procedure to adapt them. 
The next section is aimed at gaining some intuition on parameters importance in LM-CMA. 

\subsection{Sensitivity to Parameters}
\label{sectsscalingm}

\begin{figure}[!ht]
\begin{center}
	\includegraphics[width=0.788\textwidth]{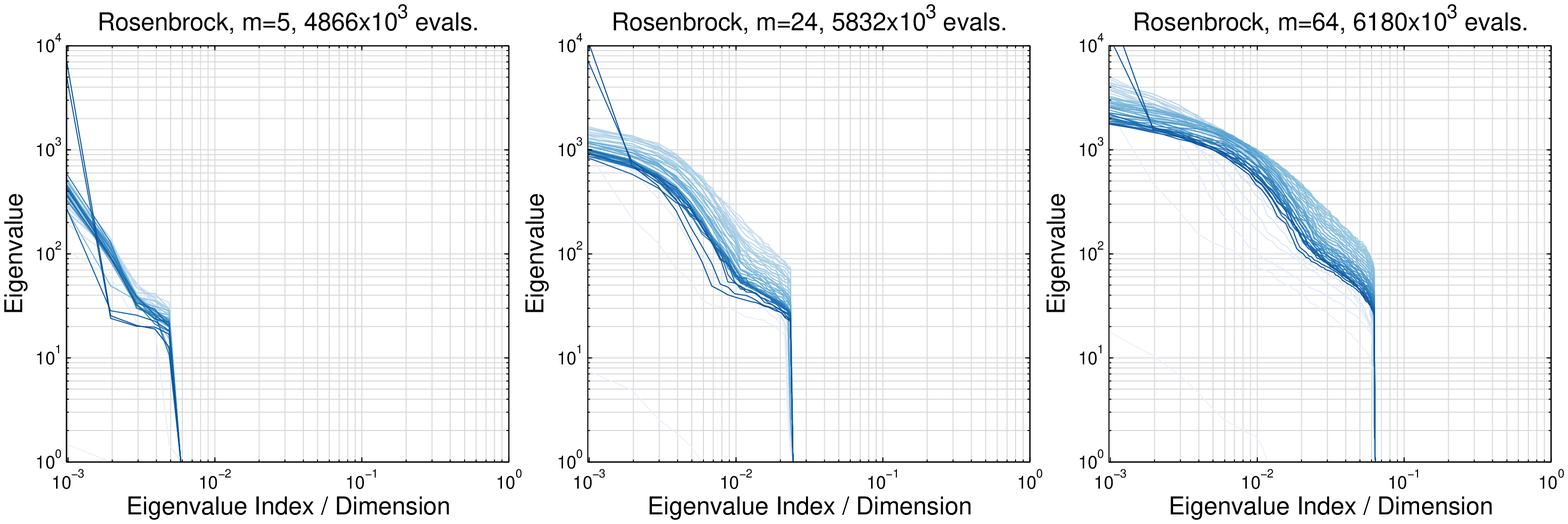}\\
  \includegraphics[width=0.788\textwidth]{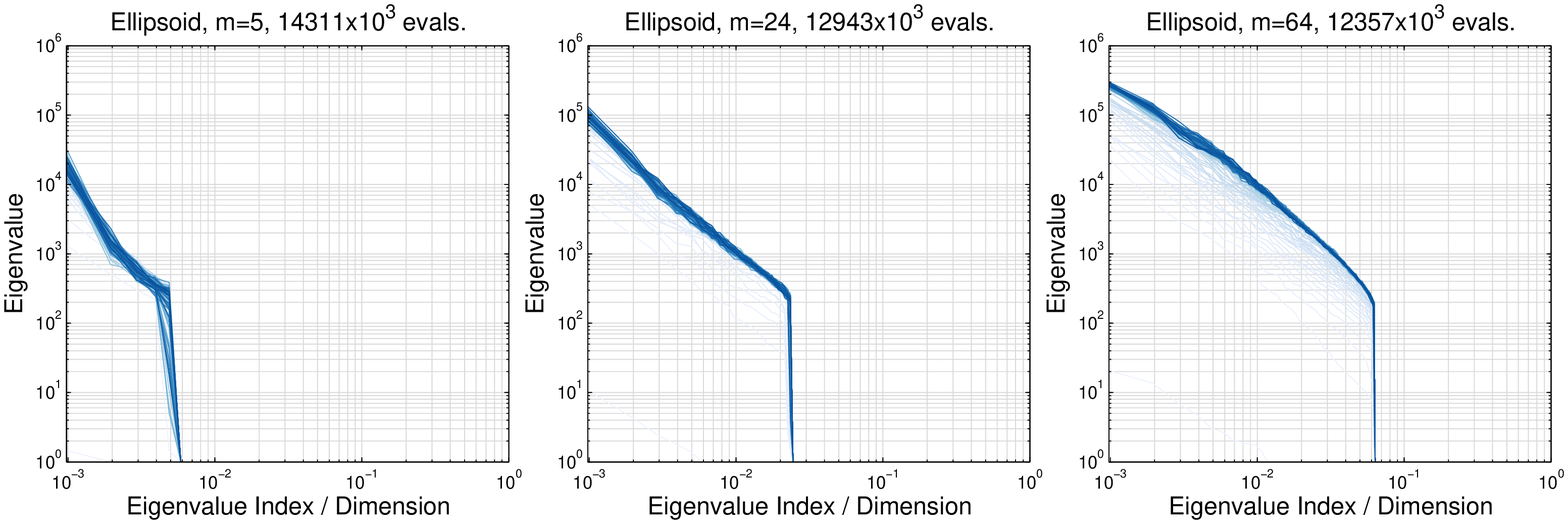}\\ 
	\includegraphics[width=0.788\textwidth]{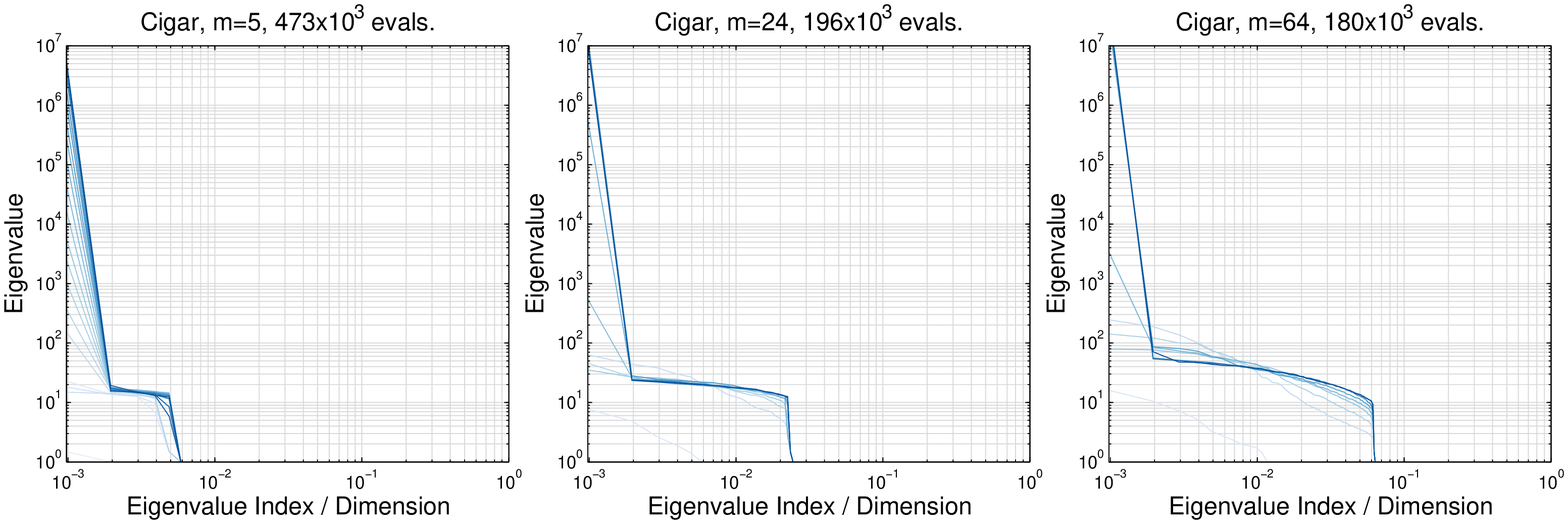}\\
  \includegraphics[width=0.788\textwidth]{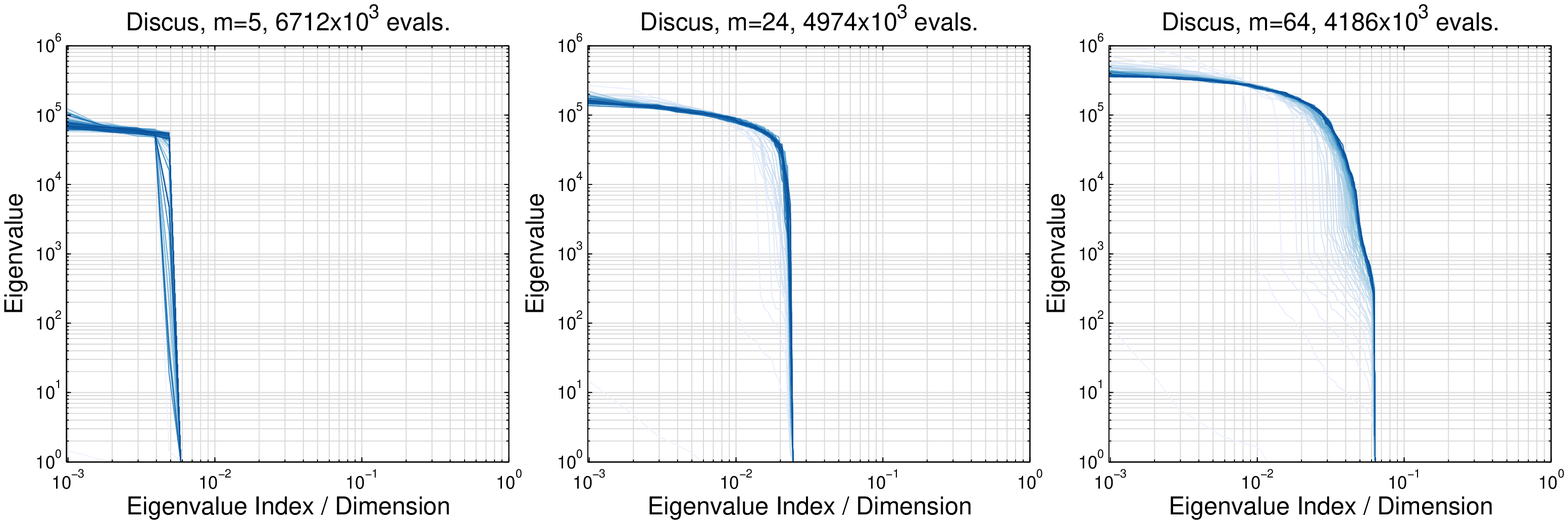}\\
	\includegraphics[width=0.788\textwidth]{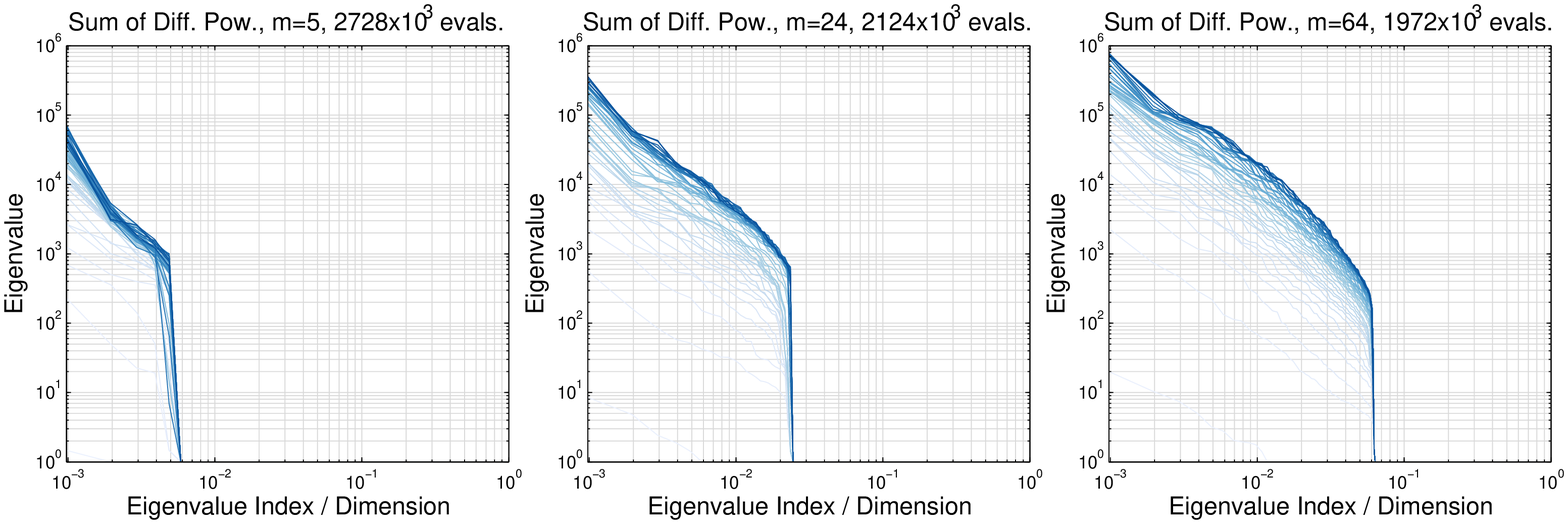}
\end{center}
\caption{\label{figeig} Eigenspectrums of $\C^t=\A^t {\A^t}^T$ for $t$ denoting iteration of LM-CMA with $m$ direction vectors ($m=5$, $m=4+\left\lfloor 3\ln 1024\right\rfloor=24$, $m=\left\lfloor 2\sqrt{n}\right\rfloor=64$) on 1024-dimensional problems. Darker (blue) lines correspond to later iterations. The number of function evaluations to reach $f(\vc{x})=10^{-10}$ is given in the title of each sub-figure.}
\end{figure}

The black-box scenario implies that the optimization problem at hand is not known, 
it is therefore hard if even possible to suggest a "right" parametrization of our algorithm that works best on all problems. 
Offline tuning in large scale optimization is also computationally expensive. 
It is rather optimistic to believe that one always can afford enough computational resources to run algorithms till the optimum on very large real-world optimization problems. Nevertheless, we tend to focus on this scenario in order to gain an understanding about scalability on benchmark problems. 

Our experience with parameter selection by exclusion of non-viable settings suggests that there exists a 
 dependency between the population size $\lambda$, number of stored vectors $m$, 
the target temporal distance between them $N_{steps}$, the learning rate $c_c$ for the evolution path and learning rate $c_1$ for the Cholesky factor update. The main reason for this is that all of them impact how well the intrinsic coordinate system defined by the Cholesky factor reflects the current optimization landscape.   
A posteriori, if $m\ll n$, it seems reasonable to store vectors with a temporal distance in order of $N_{steps}=n$ on problems where a global coordinate system is expected to be constant, e.g., on a class of problems described by the general ellipsoid model \citep{beyerconvergence}. The learning rate for the evolution path is related to both $m$ and $n$, here, we set it to $c_c = \frac{0.5}{\sqrt{n}}$ which is roughly inversely proportional to the (if affordable) suggested $m=\lfloor2\sqrt{n}\rfloor$. We found that the chosen $c_c$ is still valid for the default $m = 4 + \lfloor 3 \ln \, n  \rfloor $. We do not have a good interpretation for the learning rate $c_1 = \frac{1}{10\ln(n+1)}$. In general, we are not in favor of strongly arguing for some parameters settings against the others since as already mentioned above they are problem-dependent. A more appropriate approach would be to perform online adaptation of hyper-parameters as implemented for the original CMA-ES by \cite{loshchilov2014maximum}.

We present an analysis for $m$ which directly affects the amount of memory required to run the algorithm, and, 
thus, is of special interest since the user might be restricted in memory on very large scale optimization problems with $n>10^6$. Figure \ref{figMsens} shows that the greater the $m$ the better the performance. 
The results obtained for the default $m = 4 + \lfloor 3 \ln \, n  \rfloor $, i.e., the results demonstrated in the previous sections can be improved with $m=\lfloor2\sqrt{n}\rfloor$. The improvements are especially pronounced on \discus\ functions, where the factor is increasing with $n$ and the overall cost to solve the function reaches the one extrapolated for aCMA at $n=8192$ (see Figure \ref{figvsdim}). It is surprising to observe that $m=5$ and even $m=2$ are sufficient to solve \elli, \discus\ and \diffpow. The latter is not the case for \cigar, where small values of $m$ lead to an almost quadratic growth of run-time. The overall conclusion would be that on certain problems the choice of $m$ is not critical, while greater values of $m$ are preferable in general.

We investigated the eigenspectrum of the covariance matrix $\C^t$ constructed as $\A^t {\A^t}^T$ from the Cholesky factor $\A^t$. The results for single runs on different 1024-dimensional functions and for different $m$ are shown in Figure \ref{figeig}. 
The evolution of the eigenspectrum during the run is shown by gradually darkening (blue) lines with increasing $t$. Clearly, the number of eigenvalues is determined by $m$. The profiles, e.g., the one of \cigar, also reflect the structure of the problems (see Table \ref{table:TableFunc}). The greater the $m$, the greater condition number can be captured by the intrinsic coordinate system as can be see for \elli, \discus\ and \diffpow, that in turn leads to a better performance. However, this is not always the case as can be seen for \rosen\ that again demonstrates that optimal hyper-parameter settings are problem-dependent.

\section{Conclusions}
\label{conclusion}

We adapt an idea from derivative-based optimization to extend best performing evolutionary algorithms such CMA-ES to large scale optimization. This allows to reduce the cost of optimization in terms of time by a factor of $n/10$ and memory by a factor between $\sqrt{n}$ and $n$.  Importantly, it also often reduces the number of function evaluations required to find the optimum. 
The idea to store a limited number of vectors and use them to adapt an intrinsic coordinate system is not the only but one of probably very few ways to efficiently search in large scale continuous domains.
 We propose two quite similar alternatives: i) the storage of points and a later estimation of descent directions from differences of these points, and ii) the use of a reduced matrix $m \times n$ as in \citep{knight2007reducing} but with a modified sampling procedure to obtain linear time complexity as proposed for the Adaptive Coordinate Descent by \cite{loshchilov2013surrogate}. 

The use of the Population Success Rule is rather optional and alternative step-size adaptation procedures can be applied. 
However, we find its similarity with the 1/5-th rule quite interesting. The procedure does not make any assumption about the sampling distribution, this allowed to use the Rademacher distribution. When $n$ is large, the sampling from a $n$-dimensional Rademacher distribution resembles 
the sampling from a $n$-dimensional Gaussian distribution since the probability mass of the latter is concentrated in a thin annulus of width $O(1)$ at radius $\sqrt{n}$. 

The presented comparison shows that LM-CMA outperforms other evolutionary algorithms and is comparable to L-BFGS on non-trivial large scale optimization problems when the black-box (derivative-free) scenario is considered. Clearly, the black-box scenario is a pessimistic scenario but a substantial part of works that use finite difference methods for optimization deal with this scenario, and, thus, can consider LM-CMA as an alternative.
Importantly, LM-CMA is invariant to rank-preserving transformations of the objective function and therefore is potentially more robust than L-BFGS. The results shown in Figure 7 suggest that the use of a smaller number of direction vectors $m$ can be still efficient, 
i.e., more efficient algorithms, e.g., with adaptive $m$ (or an adaptive $m \times n$ transformation matrix) can be designed. 
It seems both promising and feasible to extend the algorithm to constrained, noisy and/or multi-objective optimization, the domains, which are both hardly accessible for L-BFGS and keenly demanded by practitioners. 
As an important contribution to the success in this direction, it would be helpful to implement online adaptation of internal hyper-parameters as already implemented in the original CMA-ES \citep{2014LoshchilovLMCMA}. 
This would ensure an additional level of invariance and robustness on large scale black-box optimization problems.

\acknowledgments

I am grateful to Mich\`ele Sebag and Marc Schoenauer for many valuable discussions and insights. 
I also would like to thank Oswin Krause, Youhei Akimoto and the anonymous reviewers 
whose interest and valuable comments helped to improve this work. 

\small
\bibliographystyle{apalike}
\bibliography{mybib}

\end{document}